\documentclass[letterpaper, 10 pt, conference]{ieeeconf}  

\IEEEoverridecommandlockouts                              

\usepackage{amsmath,amsfonts,amssymb}
\usepackage{tikz,tkz-euclide}
\usetikzlibrary{arrows,calc,shapes,positioning}
\tikzset{>=latex}
\usepackage{color,soul}
\usepackage{graphicx}
\usepackage[font=footnotesize]{subfig}
\usepackage{pgfplots}
\usepackage{caption}
\usepackage{multirow}
\usepackage[colorinlistoftodos]{todonotes}

\usepackage{changes}

\usepackage{svg}
\graphicspath{ {images/} }

\usepackage{flushend} 

\usepackage[ruled, noend]{algorithm2e}

\usepackage{multirow}

\usepackage{cite}

\usepackage{cleveref}

\title{\LARGE \bf
Gaussian Process Gradient Maps for Loop-Closure Detection in Unstructured Planetary Environments
}

\author{Cedric Le Gentil$^{1\dagger}$, Mallikarjuna Vayugundla$^{2\dagger}$, Riccardo Giubilato$^{2}$, Wolfgang St{\"u}rzl$^{2}$,\\ Teresa Vidal-Calleja$^{1}$, Rudolph Triebel$^{2,3}$
\thanks{$^{\dagger}$
	{\tt\small *The authors assert equal contribution and joint first authorship.}}%
\thanks{$^{1}$Centre for Autonomous Systems at the Faculty of Engineering and IT, University  of  Technology Sydney
       \{cedric.legentil@student.uts.edu.au, teresa.vidalcalleja@uts.edu.au\}}%
\thanks{$^{2}$German Aerospace Center (DLR), Institute of Robotics and Mechatronics
        \{firstname.lastname@dlr.de\}}%
\thanks{$^{3}$Technical University of Munich (TUM), Department of Computer Science
        \{rudolph.triebel@in.tum.de\}}%
\thanks{\textcopyright 2020 IEEE. Personal use of this material is permitted. Permission from IEEE must be obtained for all other uses, in any current or future media, including reprinting/republishing this material for advertising or promotional purposes, creating new collective works, for resale or redistribution to servers or lists, or reuse of any copyrighted component of this work in other works
}
}

\begin{document}

\bstctlcite{BibControl}

\maketitle
\thispagestyle{empty}
\pagestyle{empty}

\def\nbpoints{N}
\def\i{i}
\def\m{m}
\def\nbsubmaps{M}
\newcommand\submap[1]{\mathcal{S}^{#1}}
\newcommand\point[2]{\mathbf{p}_{#2}^{#1}}
\newcommand\x[2]{x_{#2}^{#1}}
\newcommand\y[2]{y_{#2}^{#1}}
\newcommand\z[2]{z_{#2}^{#1}}
\newcommand\xc[1]{x^{#1}}
\newcommand\yc[1]{y^{#1}}
\newcommand\zc[1]{z^{#1}}
\def\zmat{\mathbf{z}}
\newcommand\inputdomain[1]{\mathbf{x}^{#1}}
\newcommand\inputdomaini[2]{\mathbf{x}^{#1}_{#2}}
\def\inputdomainmat{\mathbf{X}}
\newcommand\elevationfunction[1]{f({#1})}

\begin{abstract}
The ability to recognize previously mapped locations is an essential feature for autonomous systems.
Unstructured planetary-like environments pose a major challenge to these systems due to the similarity of the terrain.
As a result, the ambiguity of the visual appearance makes state-of-the-art visual place recognition approaches less effective than in urban or man-made environments.
This paper presents a method to solve the loop closure problem using only spatial information.
The key idea is to use a novel continuous and probabilistic representations of terrain elevation maps.
Given 3D point clouds of the environment, the proposed approach exploits Gaussian Process (GP) regression with linear operators to generate continuous gradient maps of the terrain elevation information.
Traditional image registration techniques are then used to search for potential matches.
Loop closures are verified by leveraging both the spatial characteristic of the elevation maps ($SE(2)$ registration) and the probabilistic nature of the GP representation.
A submap-based localization and mapping framework is used to demonstrate the validity of the proposed approach.
The performance of this pipeline is evaluated and benchmarked using real data from a rover that is equipped with a stereo camera and navigates in challenging, unstructured planetary-like environments in Morocco and on Mt. Etna.

\end{abstract}


\section{Introduction}

Autonomous robots operating in unstructured and unknown environments require simultaneous localization and mapping (SLAM).
Given the information provided by one or multiple sensors, SLAM algorithms estimate the system's pose as well as a representation of the environment.
A key feature of these systems is the ability to recognize previously mapped places, allowing for the correction of the drift inherently occurring in open-loop estimation frameworks.
Place recognition or, more specifically, loop closure detection is a well-studied topic in robotics.
The development of appearance-based 2D local feature detectors and descriptors has enabled many reliable and large-scale place recognition systems using vision~\cite{LowryTRO16}.

\begin{figure}
	\centering
    \def\rowheight{1.46cm}
    \begin{tabular}{c c c c}
        \includegraphics[height=\rowheight]{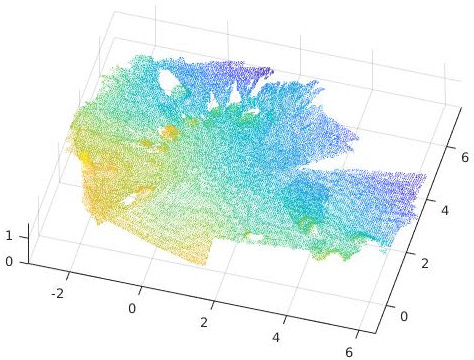}
        &
        \includegraphics[height=\rowheight]{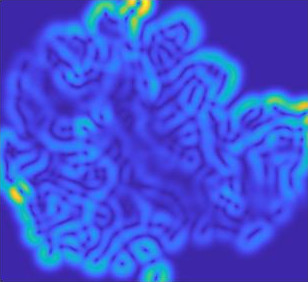}
        &
        \includegraphics[height=\rowheight]{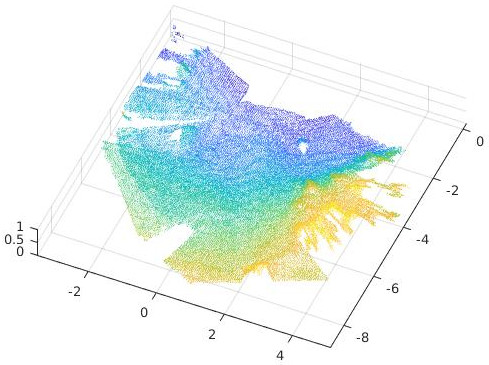}
        &
        \includegraphics[height=\rowheight]{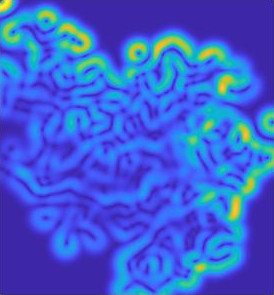}
        \\
        (a) & (b) & (c) & (d)
        \\
        \\
    \end{tabular}
    \def\rowheight{3.1cm} 
    \begin{tabular}{c c}
	    \includegraphics[height=\rowheight]{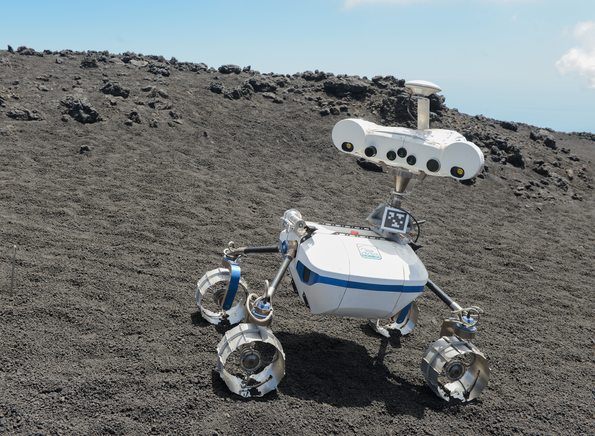} \label{fig:lru}
        &
	    \includegraphics[height=\rowheight]{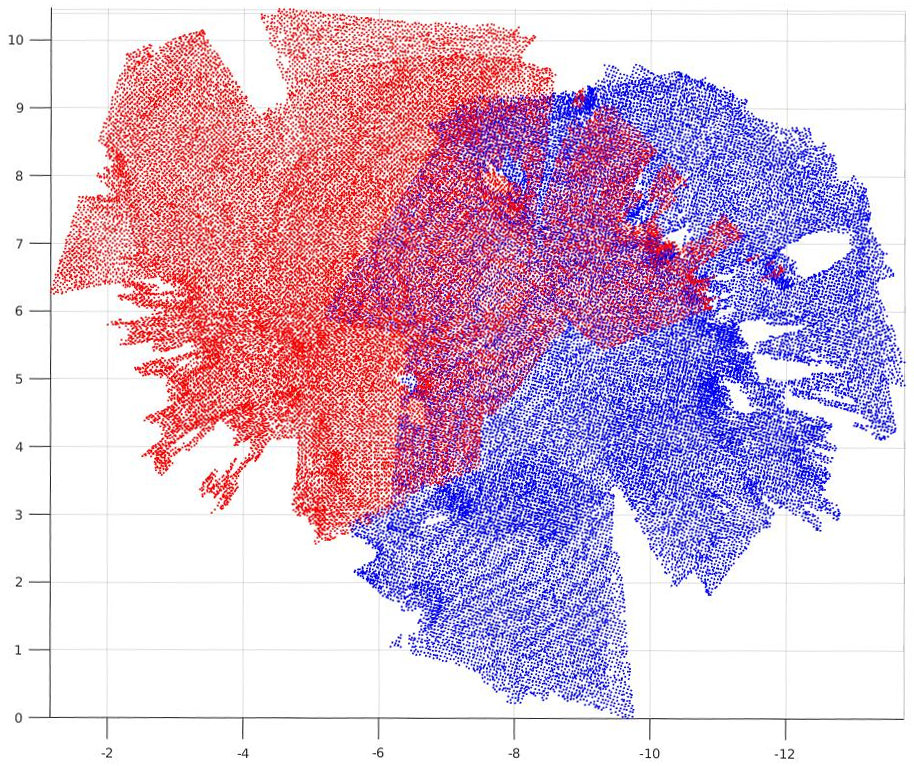}
        \\
        (e) & (f)
    \end{tabular}
    \caption{The proposed method performs loop-closure detection in challenging unstructured scenarios based on geometric representations of the environment. The input maps (a) and (c) are used to generate Gaussian gradient maps of the terrain elevation (b) and (d), respectively. These gradient maps are matched via a feature-based RANSAC algorithm. The platform used in our experiments on Mt.\ Etna (Italy) is shown in (e), photo courtesy of Esther Horvath. (f) shows the successful loop-closure detection from (b) and (d) verified from aligning input pointclouds using their poses.}
	\label{fig:lru}
\end{figure}

In this paper, we are interested in tackling the place recognition problem encountered in planetary exploration.
Unstructured planetary alike environments make the place recognition task extremely challenging using vision-based systems, especially when viewpoints of the same location differ due to the exploratory nature of the robot's path.
In the literature, besides appearance, the 3D structure of the environment has also been exploited for loop closure detection.
Features extracted from registered 3D point clouds coming from stereo-systems, depth cameras or LiDARs are commonly used in the recognition tasks.
This 3D representation, however, is usually sparse, noisy and incomplete.
Moreover, the existing 3D descriptors are in general not as reliable as the visual features, but more importantly for our application, are also challenged by the lack of structure of the terrain.

The hereby proposed method aims to establish correspondences for loop closure detection by leveraging the information contained in the 3D representation of the environment combined with 2D robust visual descriptors.
Based on gravity-aligned 3D point clouds of the robot's vicinity, our approach introduces a novel continuous and probabilistic representation of the terrain elevation and its variations.
This new type of map can be sampled into an image-like data structure, and therefore, be used to perform visual feature extraction and matching.


The proposed approach makes use of Gaussian Process (GP) regression~\cite{Rasmussen2006} in association with linear operators~\cite{Sarkka2011} to build continuous representations of the gradient of the terrain elevation.
These high entropy representations, denoted by GP gradient maps, deal with both data uncertainty and incompleteness effectively thanks to the continuous and probabilistic nature of GPs.
Our approach takes advantage of this high entropy gradient maps to extract visual-like features and descriptors to find potential correspondences with previously mapped areas.
Given these correspondences, a RANSAC-based $SE(2)$ registration approach on the GP gradient maps and the associated uncertainties are used to verify the loop closure detection.
A stereo-visual-inertial SLAM system is used to demonstrate the validity and performance of the proposed approach in an arid-desert and a volcano area, emulating the challenges encountered in planetary exploration scenarios.

In summary, the paper contributions are:
\begin{itemize}
\item A novel loop closure detection pipeline for unstructured planetary environments using GP and linear operators for point-cloud-based mapping systems.
\item A representation based on continuous and probabilistic gradient maps that enables both, 2D feature matching and SE(2) registration in a RANSAC-based approach that makes use of the uncertainty information to robustly detect loop closures.
\item An experimental validation of the proposed approach in a submap-based SLAM system with real datasets collected on Mt. Etna and in Morocco.
\end{itemize}
Fig.~\ref{fig:lru} shows an example of the input and output data from our proposed loop-closure detection approach using data from Mt. Etna.

The remainder of the paper is organized as follows: in Section \ref{sec:relatedWork}, a short overview of the existing work on place recognition with a focus on point-cloud-based methods as well as on the usage of GPs for modelling of the environment is presented. In Section~\ref{sec:methodOverview}, the overview of the proposed method is described, followed by the sections~\ref{sec:gp_maps} and \ref{sec:loopClosure} where the two main components of the pipeline, i.e. the GP gradient map representation and the loop closure validation are explained. Section~\ref{sec:implementation} describes briefly the submap based SLAM system, which we used to validate the proposed method along with the implementation
details regarding the Gaussian gradient maps. Finally, the experiments and results are presented in Section~\ref{sec:experiments} with some concluding remarks in Section~\ref{sec:conclusions}.


\section{Related Work}\label{sec:relatedWork}
Establishing loop closures in visual SLAM is traditionally performed by detecting the similarity of the current image to a database.
Visual similarity is usually expressed as the co-occurrence of image features such as SIFT~\cite{lowe2004distinctive}, SURF~\cite{bay2006surf} or ORB~\cite{rublee2011orb}, often detected by means of aggregation techniques such as bag-of-words~\cite{GalvezTRO12}, VLAD~\cite{arandjelovic2013all} or the more recent HBST~\cite{2018-schlegel-hbst} targeted at binary descriptors.
Pure image-based approaches suffer to relocalize while revisiting the same places from different viewpoints and lighting conditions.
For this reason, many approaches to loop closure detection leverage the 3D structure of the environment in the form of point clouds, obtained from LiDAR sensors, stereo cameras or dense multi-view stereo from monocular observations.
The authors of~\cite{gawel20173d} evaluate a localization system where LiDAR point clouds captured by a ground vehicle are matched to point clouds from dense visual structure-from-motion through a variety of 3D descriptors.
In~\cite{gawel2016structure} global localization of a camera in LiDAR maps is obtained by computing and matching structural descriptors, such as the 3D Gestalt~\cite{bosse2013place}, on both the map from visual SLAM and from the aggregated LiDAR point clouds.
Similarly, monocular point clouds are matched to a global LiDAR map in~\cite{caselitz2016monocular} for the purpose of online localization although requiring initial pose priors.

Loop closure detection from homogeneous data (point clouds to point clouds) is addressed in several ways.
For LiDAR scans, relocalization can be achieved using global descriptors such as ScanContext~\cite{kim2018scan} where the spatial distribution of points is encoded in an image, which is matched to a database.
The approach presented in~\cite{guo2019local} relies on matching local 3D features, derived from SHOT~\cite{Tombari_2011}, followed by a probabilistic voting step to reject false transformation hypotheses.
In~\cite{Dube-RSS-18,dube2020segmap,Dube2018}, LiDAR clouds are discretized in segments, representing clusters of points spatially separated between each other.
In the latest iteration of the author's pipeline, segments are encoded into low dimensional descriptors from an auto-encoder and matched to a database to either close loops or relocalize on a prior map.
Recently, deep learning techniques have been leveraged for point cloud registration to either obtain more accurate local descriptors~\cite{yew20183dfeat,gojcic2018perfect} or robust keypoints~\cite{wang2018learning}.
Although demonstrating higher performances than handcrafted methods, the computational complexity makes their implementation on resource-constrained vehicles challenging.


Although not for loop closure detection, application of Gaussian Processes to produce continuous and probabilistic elevation maps from 3D point clouds has been previously studied in the literature.
In~\cite{Vasudevan2009Large}, GPs have been used to model the spatial correlation of observed terrain data to infer topography from the new observations.
GP elevation maps in~\cite{Liye15fusion} are used as a prior for Bayesian fusion. Other GP-based mapping properties such as occupancy~\cite{OCallaghanIJRR12}, thickness~\cite{Teresa2013ICRA}, implicit surface~\cite{WilliamsGPIS,WuRAL2020}, among others, have been extensively studied in recent years.
Most of these works exploit the probabilistic nature and the inference capabilities to fill-up areas of missing data, to produce maps of a desirable resolution, to filter noise or for data fusion.
Our work, in contrast, exploits the derivative (through linear operators) of the continuous elevation terrain to extract features of the gradient maps and the probabilistic aspect to limit the area of interest and validate loop closures.

Moreover, the use of GP regression and linear operators has already been leveraged for robotics state estimation in~\cite{LeGentil2020} to create continuous and accurate pre-integrated measurements from noisy inertial readings.
In~\cite{MartensRAL16}, GP regression is used to recover implicit surfaces with normal estimates for 3D surface reconstruction.
In a similar way, here, it is used to generate gradient maps.
Operating on the gradient instead of elevation directly makes the proposed pipeline elevation-invariant, a key aspect that is leveraged for registration in the loop closure validation process.

\section{Method overview}\label{sec:methodOverview}

\begin{figure}

    \centering
    \begin{tikzpicture}[auto]
        \tikzstyle{block} = [draw, fill=white, rectangle, minimum height = 4.2em, text width = 6em,  minimum width = 6em, align = center, node distance = 5em]
        \def\inputdist{3em}

        \coordinate (new_data){};
        \node [block, right = 2.0em of new_data, execute at begin node=\setlength{\baselineskip}{9pt}] (gp_map){\footnotesize Gaussian elevation gradient map inference};
        \node [block, right = \inputdist of gp_map, execute at begin node=\setlength{\baselineskip}{9pt}] (loop_detection){\footnotesize Loop-closure detection \textit{\scriptsize \color{black}{Feature-based RANSAC matching}}};
        \node [below = 3.1em of new_data, text width = 8em, anchor = west, align =center, execute at begin node=\setlength{\baselineskip}{8pt} ] (previous_map){\color{darkgray} \scriptsize Gaussian gradient map of previously explored areas};
        \coordinate [below = 3.1em of new_data] (coordinate_previous_map){};
        \node [right = 4.5em of loop_detection] (out_flag){};
        \node [right = 0em of loop_detection, anchor= north west , text width = 4.5em, align = center, execute at begin node=\setlength{\baselineskip}{8pt} ] (so2){\scriptsize \color{darkgray} and $SE(2)$ transformation};
        \node [right = 0em of loop_detection, anchor= south west , text width = 4.5em, align = center , execute at begin node=\setlength{\baselineskip}{8pt}] (flag){\color{darkgray}\scriptsize Loop closure: false/true};

        \draw[->] (new_data) -- node[text width = 3em, align = center, execute at begin node=\setlength{\baselineskip}{8pt}]{\color{darkgray}\scriptsize Point cloud} (gp_map);
        \draw[->] (gp_map) -- node[text width = 3em, align = center, execute at begin node=\setlength{\baselineskip}{8pt}]{\color{darkgray}\scriptsize GP gradient} (loop_detection);
        \draw[->] (loop_detection) -- (out_flag);
        \draw[->] (coordinate_previous_map.west) -| (loop_detection.south);


    \end{tikzpicture}
	\caption{
        Overview of the proposed pipeline of place recognition for loop-closure detection based on geometric representations of the environment.
	}
	\label{figure:overview}
\end{figure}
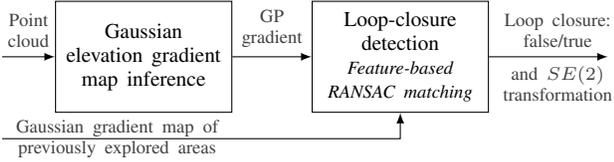

The proposed method performs loop-closure detection given gravity-aligned 3D point clouds of the environment.
As shown in Fig.~\ref{figure:overview}, our method is divided into two distinct parts.

The first one is the application of GP regression~\cite{Rasmussen2006} and linear operators on kernels~\cite{Sarkka2011} to convert incoming point clouds into Gaussian gradient maps of the terrain elevation.
It relies on the assumption that the true underlying geometry can be modelled with terrain elevation maps.
This assumption is accurate as per the rarity of overhanging structures in the considered scenario.
The key feature of this continuous probabilistic representation is the possibility to sample the elevation's variance and gradient at any point in the $x$-$y$ plane.

The second part matches newly inferred gradient maps to previously mapped areas.
A RANSAC matching procedure is conducted upon traditional visual features extracted from the gradient maps while leveraging the geometric ($SE(2)$ transformations) and probabilistic (availability of variance) properties of the novel representation.

Note that this method is highly versatile as it is agnostic to the mapping technique used to generate the 3D-geometry of the environment.
Therefore, one can directly apply this pipeline to the output of any mapping system that generates 3D data regardless of the modality used (LiDAR, stereo-vision, RGBD cameras, etc.).
The loop closure detection in the context of this paper aims at establishing the correspondence between the current and previously mapped areas in the form of submaps.
Online loop closure triggering is out of the scope of this work.
In the following, for convenience, we consider a collection of submaps $\submap{\m}$, with $\m$ the index of the submap, to represent both the previously explored area and the newly discovered one.

\section{GP Gradient Map Representation} \label{sec:gp_maps}

This section introduces the derivations associated with the generation of the \emph{gradient maps} based on Gaussian Process regression~\cite{Rasmussen2006} and the use of linear operators on the kernel covariance function~\cite{Sarkka2011}.

\subsection{Gaussian Process regression}
Let us consider a terrain elevation submap $\submap{\m}$ constituted of $\nbpoints^{\m}$ points $\point{\m}{\i} = \begin{bmatrix}\x{\m}{\i}& \y{\m}{\i} &\z{\m}{\i}\end{bmatrix}^\top$ ($\i = 1,\cdots,\nbpoints^{\m}$) with $m$ being the index of the submap.
    The $z$-coordinate of $\point{\m}{\i}$ is represented as a function $\z{\m}{\i} = \elevationfunction{\inputdomaini{\m}{\i}}$ at the $x$ and $y$-coordinates of the same point, with $\inputdomaini{\m}{\i} = \begin{bmatrix}\x{\m}{\i}&\y{\m}{\i}\end{bmatrix}$.

\def\abscissa{\mathbf{x}}

\newcommand{\kernel}[2]{\mathbf{K}(#1,#2)}

\newcommand{\kernelfunction}[2]{k(#1,#2)}
Considering the elevation function $\zc{\m}$ modelled with a zero-mean GP~\cite{Rasmussen2006} as
\begin{align}
    &\elevationfunction{\inputdomain{\m}} \sim \mathcal{GP}\big(0,\kernelfunction{\inputdomaini{\m}{\i}}{\inputdomaini{\m}{\i'}} \big),
    \nonumber
    \\
    \z{\m}{\i} &= \elevationfunction{\inputdomaini{\m}{\i}} + \eta_{\i}, \ \ \ \ \eta_{\i} \sim \mathcal{N}(0,\sigma^2_{z}),
    \label{eq:gp_model}
\end{align}
where $\kernelfunction{\inputdomaini{\m}{\i}}{\inputdomaini{\m}{\i'}}$ is the kernel covariance function, the elevation can be inferred for any $x$-$y$ coordinates $\inputdomain{\m}$ as
\begin{align}
    \z{\m}{*}(\inputdomain{\m}) &= \kernel{\inputdomain{\m}}{\inputdomainmat} \big[ \kernel{\inputdomainmat}{\inputdomainmat} + \sigma^2_{z} \mathbf{I}\big]^{-1}\mathbf{\zmat},
    \label{eq:gp_classic_mean}
    \\
    \text{var}(\z{\m}{*}) &= \kernel{\inputdomain{\m}}{\inputdomain{\m}}
    \nonumber
    \\
    - &\kernel{\inputdomain{\m}}{\inputdomainmat} \big[\kernel{\inputdomainmat}{\inputdomainmat} + \sigma^2_{z} \mathbf{I}\big]^{-1} \kernel{\inputdomain{\m}}{\inputdomainmat}^\top,
    \label{eq:gp_classic_variance}
\end{align}
with $\inputdomainmat$ the matrix built by stacking the $x$-$y$ coordinates $\inputdomain{\m}$ of the training points, $\zmat$ the vector of training values $\z{\m}{\i}$ at $\inputdomainmat$, and $\kernel{.}{.}$ the matrix of covariances evaluated with the kernel covariance function $\kernelfunction{.}{.}$ between each pair of arguments.

\subsection{Gaussian gradient maps}

\newcommand\linearoperator[1]{\mathcal{L}_{#1}}

In order to propose a method that is invariant with respect to the elevation of the submaps' origin, the proposed framework generates a representation of the gradient of the submaps' elevation.
Overall this helps the feature matching robustness, and allows for the use of difference-based metrics to register gradient maps together.
The $x$ and $y$ gradients of $\zc{\m}$ are directly inferred from the GP model \eqref{eq:gp_model} by applying the differentiation linear operator to the covariance kernel as introduced in~\cite{Sarkka2011}:
\begin{align}
    \frac{\partial \zc{\m}}{\partial \xc{\m}} = \frac{\partial \kernel{\inputdomain{\m}}{\inputdomainmat}}{\partial \xc{\m}} \big[ \kernel{\inputdomainmat}{\inputdomainmat} + \sigma^2_{z} \mathbf{I}\big]^{-1}\mathbf{\zmat},
    \label{eq:gp_operator_x}
    \\
    \frac{\partial \zc{\m}}{\partial \yc{\m}} = \frac{\partial \kernel{\inputdomain{\m}}{\inputdomainmat}}{\partial \yc{\m}} \big[ \kernel{\inputdomainmat}{\inputdomainmat} + \sigma^2_{z} \mathbf{I}\big]^{-1}\mathbf{\zmat}.
    \label{eq:gp_operator_y}
\end{align}

\begin{figure}
    \centering
    \newcolumntype{C}[1]{>{\centering\let\newline\\\arraybackslash\hspace{0pt}}m{#1}}
    \def\colsize{2.5cm}
    \begin{tabular}{C{\colsize} C{\colsize} C{\colsize}}
        \includegraphics[clip, width=\colsize,trim=6.6cm 11.5cm 6.5cm 11.5cm]{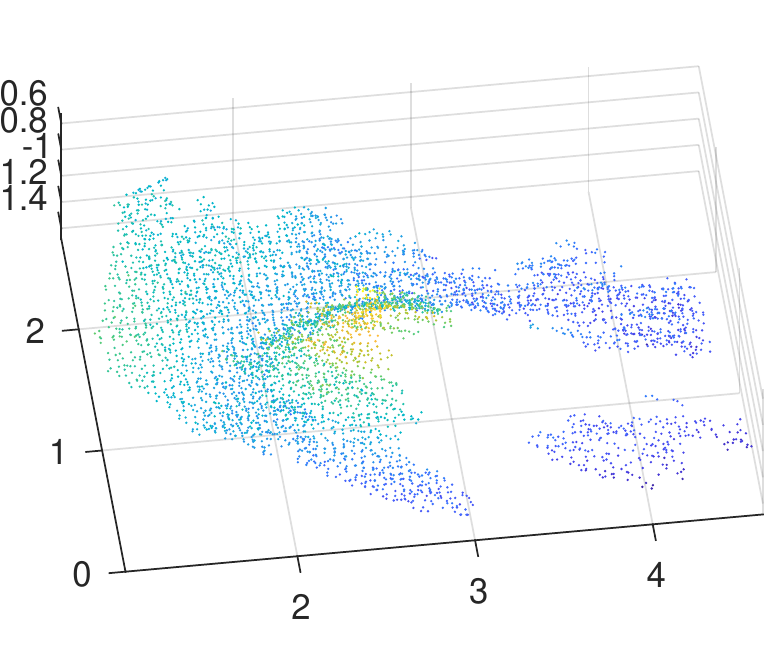}
        &
        \includegraphics[clip, width=\colsize,trim=8.9cm 13.95cm 8.9cm 12.8cm]{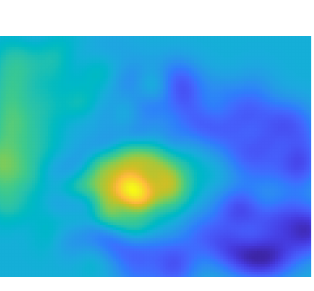}
        &
        \includegraphics[clip, width=\colsize,trim=8.9cm 13.95cm 8.9cm 12.8cm]{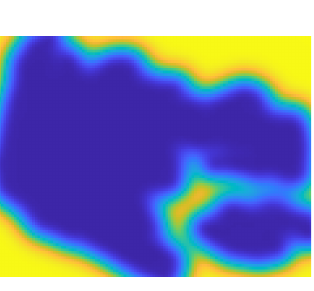}
        \\[-1ex]
        \small(a) Submap's point cloud (view A)
        &
        \small(b) Elevation GP map
        &
        \small(c) Elevation's variance
        \\
        \includegraphics[clip, width=\colsize,trim=6.6cm 13cm 6.5cm 13cm]{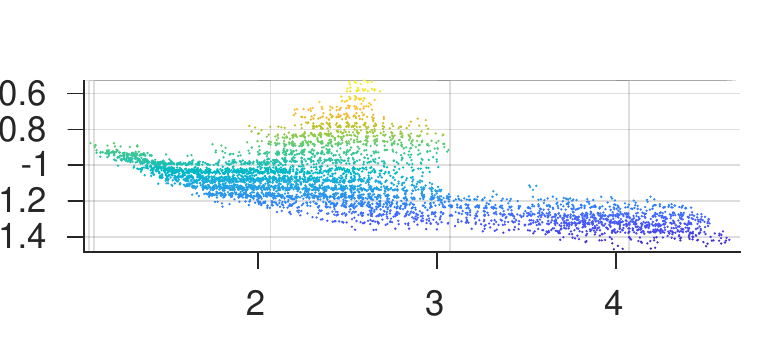}
        &
        \includegraphics[clip, width=\colsize,trim=8.9cm 13.95cm 8.9cm 12.8cm]{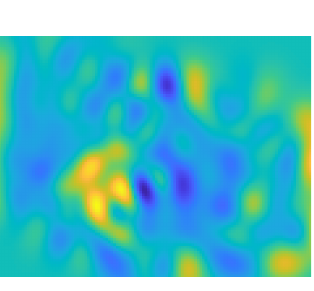}
        &
        \includegraphics[clip, width=\colsize,trim=8.9cm 13.95cm 8.9cm 12.8cm]{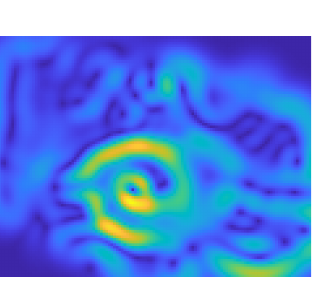}
        \\[-1ex]
        \small(d) Submap's point cloud (view B)
        &
        \small(e) Elevation's gradient along $x$
        &
        \small(f) GP gradient map
    \end{tabular}
    \caption{Example of gradient map generation from a 3D point cloud. (a) and (d) are two different viewpoints of the input 3D point cloud. (b) and (c) are the GP continuous representations of the terrain's elevation and the associated variance (from \eqref{eq:gp_classic_mean} and \eqref{eq:gp_classic_variance}). (e) represents an intermediate step (from \eqref{eq:gp_operator_x}) needed to compute the GP gradient map shown in (f).}
    \label{figure:gradient_map}
\end{figure}

From these continuous representations, the proposed method infers the elevation variance and the gradient submaps as mono-channel image-like data-structures.
For a submap $\submap{\m}$, $I_v^{\m}$ denotes the variance image-like data-structure inferred from \eqref{eq:gp_classic_variance}, and $I_g^{\m}$ the elevation's gradient computed from \eqref{eq:gp_operator_x} and \eqref{eq:gp_operator_y} using $g= \sqrt{{\frac{\partial \zc{\m}}{\partial \xc{\m}}}^2 + {\frac{\partial \zc{\m}}{\partial \yc{\m}}}^2}$.
Fig.~\ref{figure:gradient_map} shows an example of a gradient map and the associated terrain variance.
As the figure shows, gradient maps produce a higher entropy representation compared to simple elevation maps.
Thus, visual features are more present in these maps.

\section{Loop-closure detection}\label{sec:loopClosure}

\begin{figure}
	\centering
	\includegraphics[width=\linewidth]{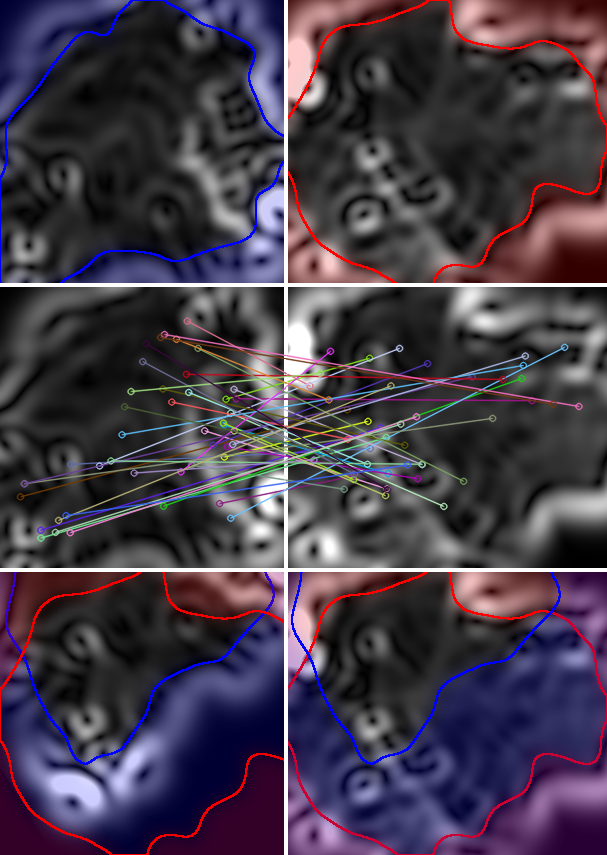}
    \caption{Visualization of the loop-closure check via gradient map matching. Top row: Input gradient submaps from Etna3 dataset (overlaid with corresponding variance in red and blue for intuitive visualization). Middle row: 2D descriptor (SURF) correspondences in low variance areas between the two gradient submaps. Bottom row: Gradient maps aligned after finding the loop closure (overlaid with both submaps' variance for intuitive visualization).}
\label{fig:loopClosureVisualisation}
\end{figure}

\def\resolution{r}
Given two instances of Gaussian gradient maps, ${I^{d}_{g}}$ and ${I^{q}_{g}}$, the proposed method classifies a pair of maps as loop-closure or not by attempting a feature-based registration between ${I^{d}_{g}}$ and ${I^{q}_{g}}$.
Note that the superscript $d$ corresponds to the index of a submap in the database of previously explored areas, and the superscript $q$ corresponds to the index of the query submap.
First, 2D keypoints are extracted individually in ${I^{d}_{g}}$ and ${I^{q}_{g}}$.
The keypoint detection leverages the gradient maps' variances, ${I^{d}_{v}}$ and ${I^{q}_{v}}$, by applying masks to reject keypoint candidates that are far from the GP training data (high variance areas).
Each keypoint is associated with a 2D descriptor computed on the gradient map instances ${I^{d}_{g}}$ and ${I^{q}_{g}}$.
As illustrated in Fig.~\ref{fig:loopClosureVisualisation}, the descriptors are matched across both gradient maps based on a brute-force strategy and $L_2$ proximity metric.

\newcommand\asso[1]{\mathfrak{a}_{#1}}
\def\forssd{\inputdomain{d,q}}
\def\assoset{\mathcal{A}}
This last step results in a set $\assoset$ of keypoint associations $\asso{i}$ across both maps.
Given this set $\assoset$, the proposed method attempts the $SE(2)$ registration of ${I^{q}_{g}}$ to ${I^{d}_{g}}$ in a RANSAC algorithm shown in Alg.~\ref{alg:loopClosureDetection}.
The different elements employed are:
\begin{itemize}
    \item \textit{SelectRandomAsso}: the random selection of a pair of associations $\asso{a}$ and $\asso{b}$ from $\assoset$.
    \item \textit{GetSE2Transform}: the estimation of a $SE(2)$ transformation $\mathbf{T}_i$, given a subset of $\assoset$, based on centroid alignment and SVD.
    \item \textit{GetInliers}: the identification of the inlier set of association $\assoset_i$, given $\assoset$ and a $SE(2)$ transformation $\mathbf{T}_i$, based on 2D distance threshold.
    \item \textit{ComputeSSDMetric}: the computation of the similarity metric $h_i$ between two registered gradient maps as the Sum of Squared gradient Differences (SSD) weighted by the inverse of the product of the corresponding variances $h_i = \underset{{\scriptstyle\forssd}}{\Sigma}\frac{(I_g^q(\mathbf{T}_i\forssd)-I_g^d(\forssd))^2}{I_v^q(\mathbf{T}_i\forssd)I_v^d(\forssd)}$.
    \item \textit{AcceptTransformation}: the decision $true/false$ to accept the estimated transformation $\mathbf{T}_i$ based on the SSD metric $h_i$ and the number of inliers found $\lvert \assoset_i \lvert$.

\end{itemize}

The output of this procedure is the maximum number of inliers found $n$ and the associated SSD similarity metric $h$.
Empirically we found that setting a threshold on the number of inliers appeared to be the most effective metric to classify two maps as loop closure.
Extensive tests have been conducted in our experiments to best define this threshold.
Note that this pipeline is agnostic to the choice of feature descriptor.
In our implementation, we use Speeded Up Robust Features (SURF) as their blob detection ability matches the smooth nature of the Gaussian gradient maps.


\begin{algorithm}
	\small
	\SetKwInOut{Input}{Input}
	\SetKwInOut{Output}{Output}
	\SetKwIF{If}{ElseIf}{Else}{if}{then}{else if}{else}{endif}
	\Input{
		\begin{itemize}
			\item ${I^{d}_{g}}$, ${I^{q}_{g}}$, ${I^{d}_{v}}$, ${I^{q}_{v}}$: Gradient maps and associated variance
            \item $\assoset$: Set of keypoint associations between ${I^{d}_{g}}$ and ${I^{q}_{g}}$
		\end{itemize}
	}
	\Output{
		\begin{itemize}
			\item ${n}$: maximum number of inliers found
            \item $h$: associated SSD metric
            \item $\mathbf{T}$: associated $SE(2)$ transformation estimate
		\end{itemize}
	}

	\ForEach{ until Max Nb Iterations}{
        $\{\asso{a}, \asso{b}\}$ = \textit{SelectRandomAsso}($\assoset$)\;
        ${\mathbf{T}_i}$ = \textit{GetSE2Transform}($\{\asso{a}, \asso{b}\}$)\;
        $\assoset_i$ = \textit{GetInliers}($\assoset$, ${\mathbf{T}_i}$)\;
        ${\mathbf{T}_i}$ = \textit{GetSE2Transform}($\assoset_i$)\;
        $h_i$ = \textit{ComputeSSDMetric}($\mathbf{T}_i$, ${I^{d}_{g}}$, ${I^{q}_{g}}$, ${I^{d}_{v}}$, ${I^{q}_{v}}$)\;

        \uIf{\textit{AcceptTransformation}($\lvert \assoset_i \lvert$, $h_i$) \textbf{and} $\lvert \assoset_i \lvert > n$}{
            $n$ = $\lvert \assoset_i \lvert$\;
            $h$ = $h_i$\;
            $\mathbf{T}$ = $\mathbf{T}_i$\;
        }
	}
    \textbf{return} ${\mathbf{T}}$, $n$, $h$\;

    \caption{Loop-closure classification based on $SE(2)$ registration between two Gaussian gradient maps.}
	\label{alg:loopClosureDetection}
\end{algorithm}


\section{Implementation}\label{sec:implementation}

This section contains a brief introduction of the mapping system used in our experiments, as well as implementation details regarding the Gaussian gradient maps and the overall computation time.

\subsection{Submap-based SLAM}\label{sec:submapSlam}

In order to evaluate the performance of our pipeline, we employ our submap-based SLAM framework \cite{doi:10.1002/rob.21812}, where gravity-aligned 3D submaps are incrementally built from aggregated stereo clouds and positioned according to visual-inertial pose estimates.
The size of submaps is bounded either by limits on the trajectory length from their respective origins or on the pose uncertainty.
This ensures that the resulting point clouds are locally accurate.
Submap origins are joined consecutively in a graph by relative visual-inertial constraints and, in the original implementation, inter-submap matches \cite{brand2014submap} establish loop closure constraints.
In this work, we evaluate the feasibility of establishing the loop closure between submaps using a GP representation instead of matching 3D keypoints using CSHOT descriptors as in \cite{brand2014submap}.

\subsection{Gaussian gradient maps and Computation}

\def\kernellength{l_k}
\def\kernelsigma{\sigma_k}
Our implementation is built upon the use of the square-exponential kernel
\begin{align}
    \kernelfunction{\inputdomaini{\m}{\i}}{\inputdomaini{\m}{\i'}} = \kernelsigma\text{exp}\Big(-\frac{(\inputdomaini{\m}{\i} - \inputdomaini{\m}{\i'})^\top(\inputdomaini{\m}{\i} - \inputdomaini{\m}{\i'})}{2\kernellength}\Big)
\end{align}
to generate the GP gradient submaps.
The hyperparameters $\kernelsigma$, $\kernellength$,  and $\sigma_{z}$ (data noise from \eqref{eq:gp_model}) are respectively initialized with the elevation's empirical variance for the submap, a manually-set length-scale (around $0.1\,\mathrm{m}$), and the typical elevation accuracy of the submapping system, before being fine-tuned as described in \cite{Rasmussen2006}.
The submaps are arbitrarily downsampled to 5000 points in order to reduce the computation time.
The query point inferences are performed with a spatial resolution of $0.03\,\mathrm{m/pixel}$ over the minimum $x$-$y$ bounding box of each point cloud.
Our current implementation computes each GP gradient map in approximately five minutes, and the RANSAC matching takes around seven minutes for a database of 14 submaps.
The computations were performed on a laptop with Intel Core i7-6700HQ CPU. 
The overall computation time can be greatly reduced, as mentioned in the future work discussed in Section~\ref{sec:conclusions}.

\section{Experiments and Discussion}\label{sec:experiments}

\begin{figure}
	\centering
	\subfloat[DLR Sensoric Unit of Planetary Exploration Rovers (SUPER)]{%
		\label{fig:first}%
		\includegraphics[height=1.15in]{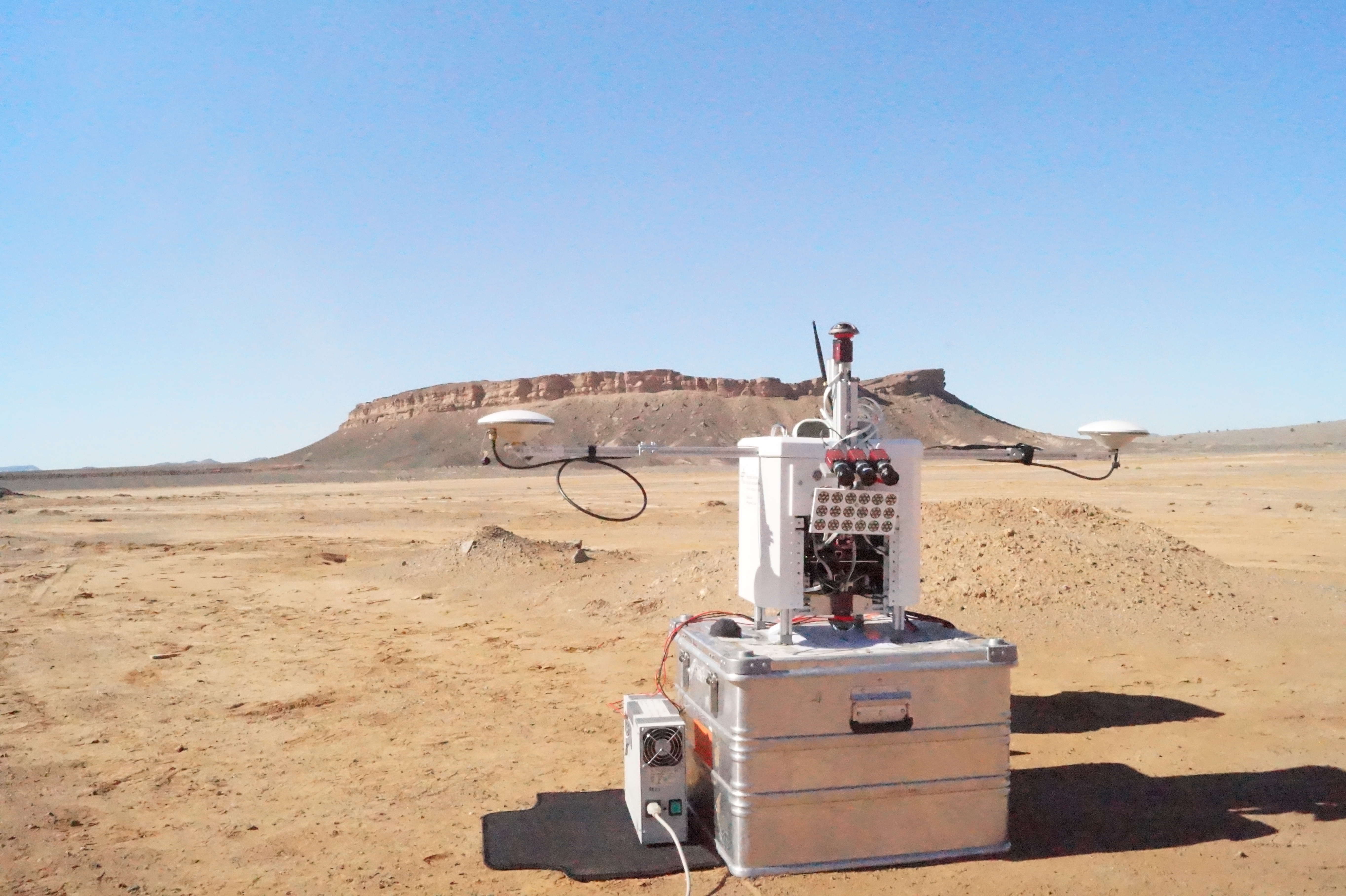}}%
	\hfill
	\subfloat[Morocco Experiment Site]{%
		\label{fig:second}%
		\includegraphics[height=1.15in]{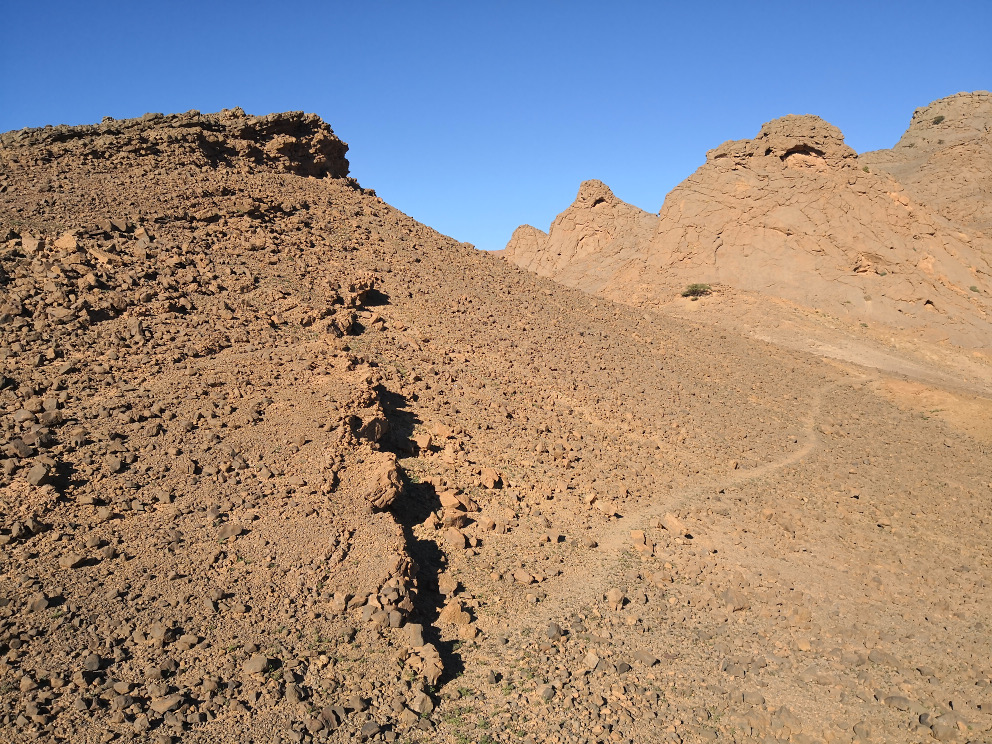}}%
	\\
	\subfloat[Etna3 Image frame]{%
		\label{fig:third}%
		\includegraphics[height=1.15in, trim=4.5cm 3cm 3.5cm 4cm, clip]{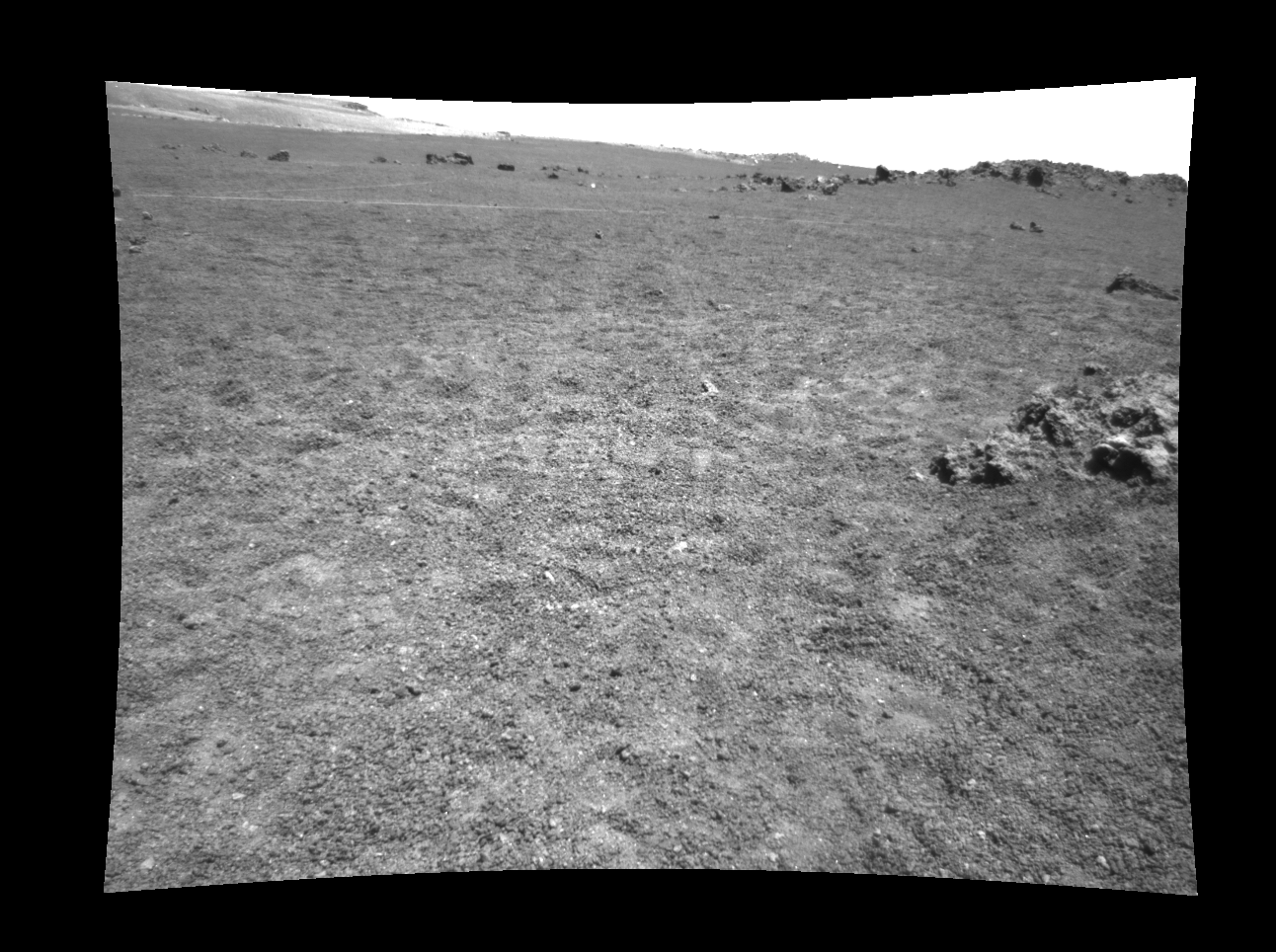}}%
	\hfill
	\subfloat[Morocco Image frame]{%
		\label{fig:fourth}%
		\includegraphics[height=1.15in]{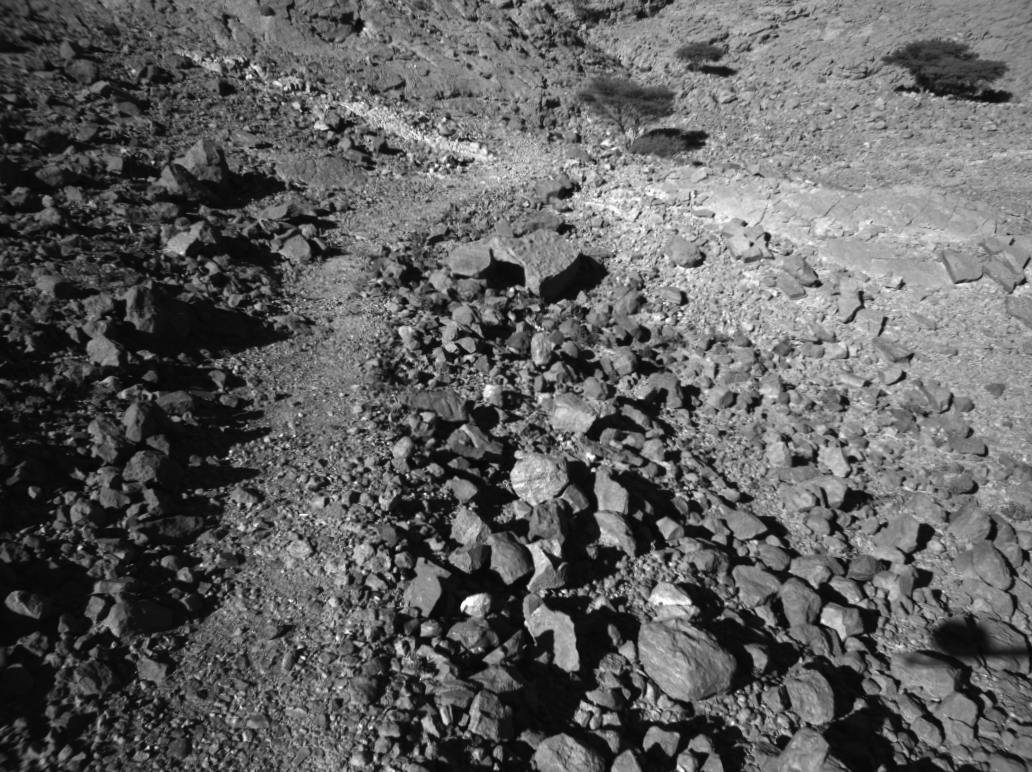}}%
	\caption{DLR Sensoric Unit
of Planetary Exploration Rovers (SUPER) at the experiment site in Morocco (a) and the location at which the particular dataset from Morocco was collected (b). Impressions of the Etna (c) and Morocco (d) landscapes as seen from the navigation camera. Notice the lack of unique visual features in (c) and the harsh lighting conditions in (d).}
	\label{fig:hcru}
\end{figure}
\begin{figure}
	\centering
	\subfloat[Etna1]{\includegraphics[width=0.49\columnwidth]{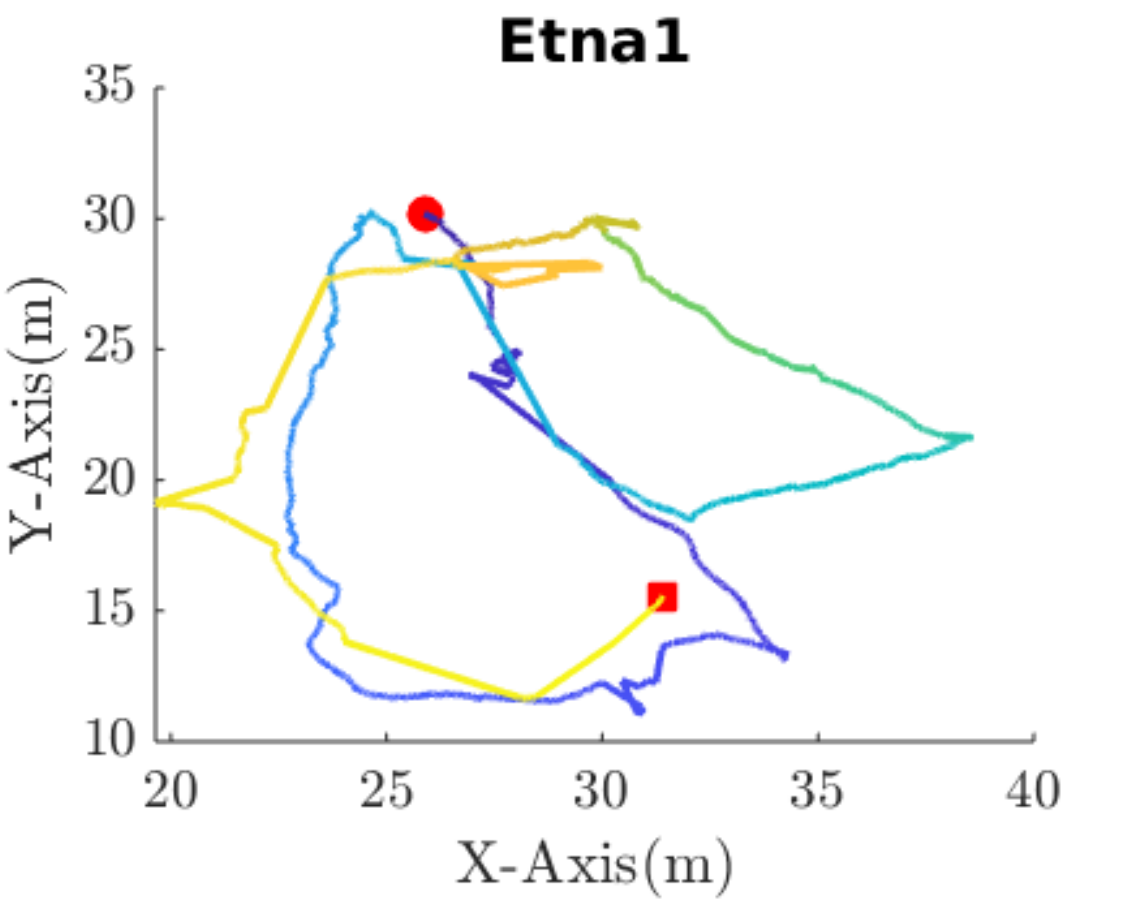}}
    \hfill
	\subfloat[Etna2]{\includegraphics[width=0.49\columnwidth]{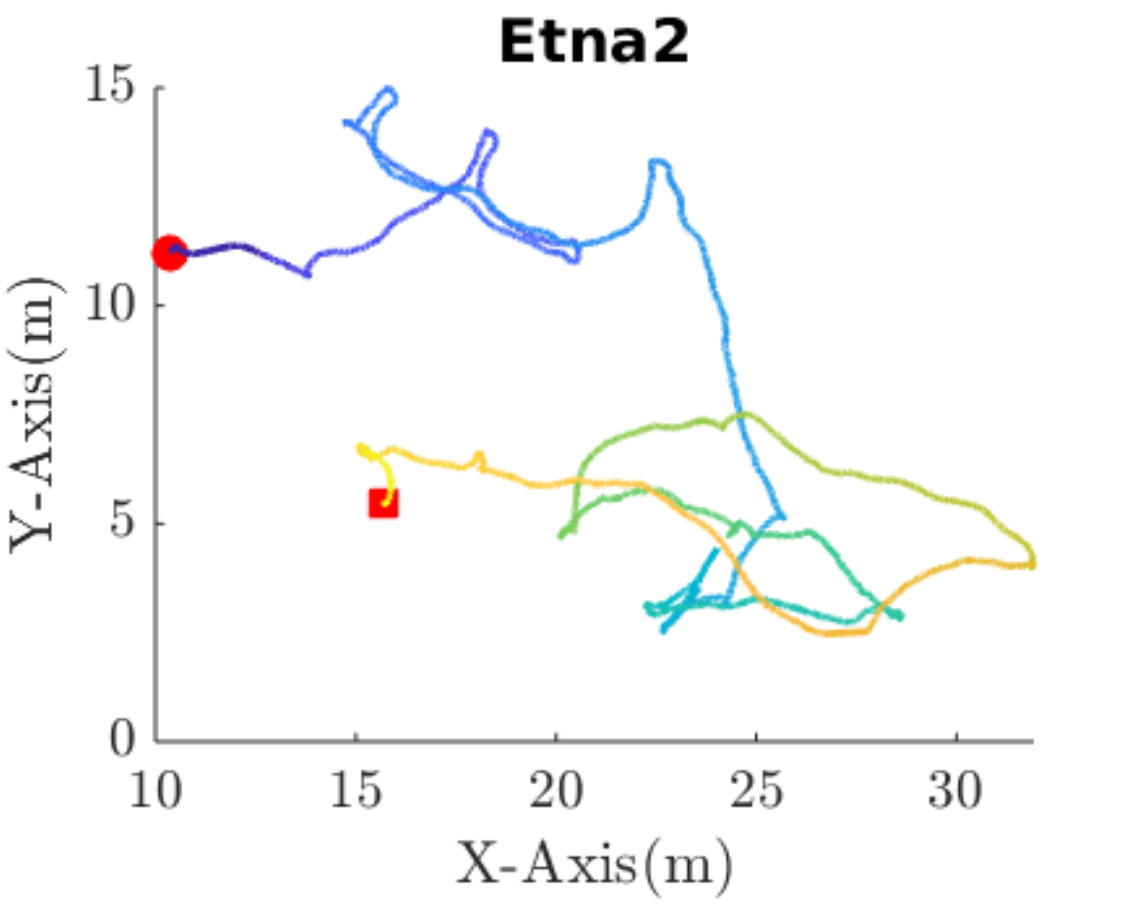}}
    \\

	\subfloat[Etna4]{\includegraphics[width=0.49\columnwidth]{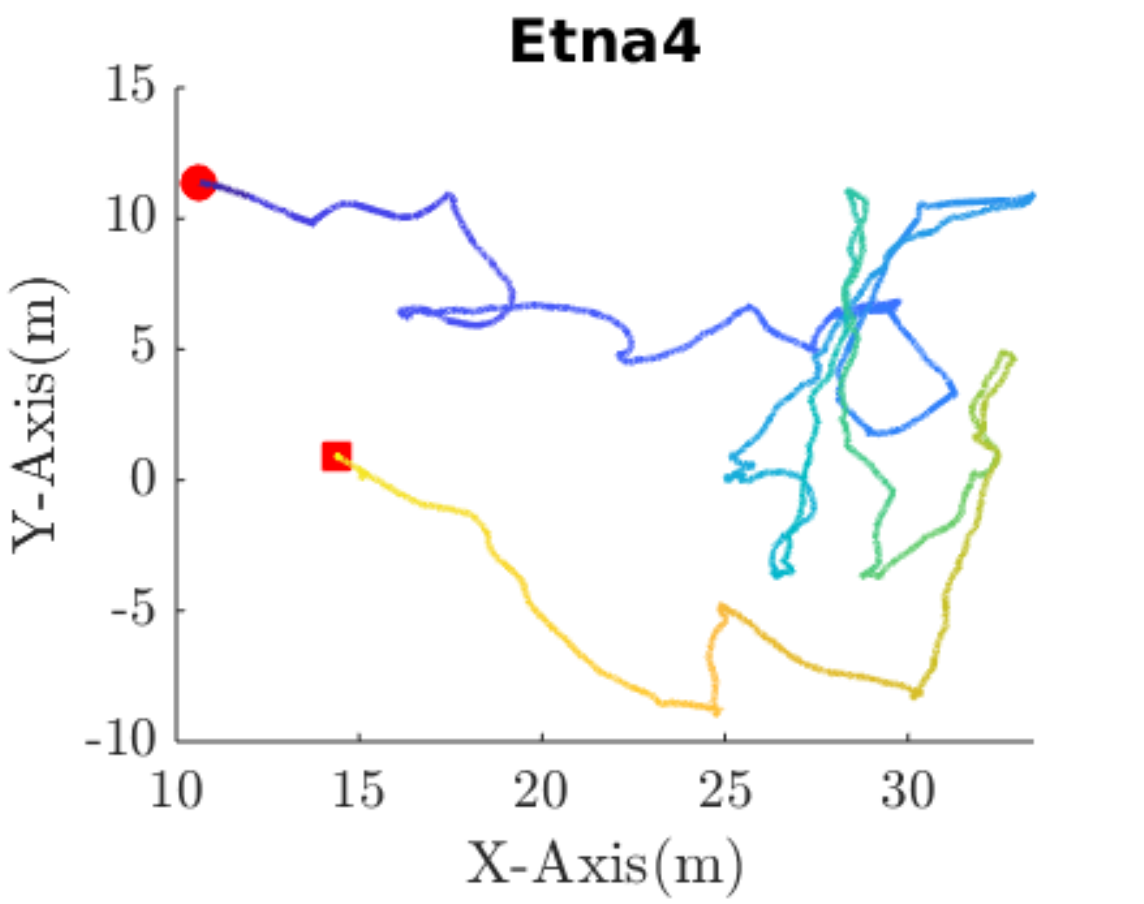}}
        \hfill
	\subfloat[Morocco]{\includegraphics[width=0.49\columnwidth]{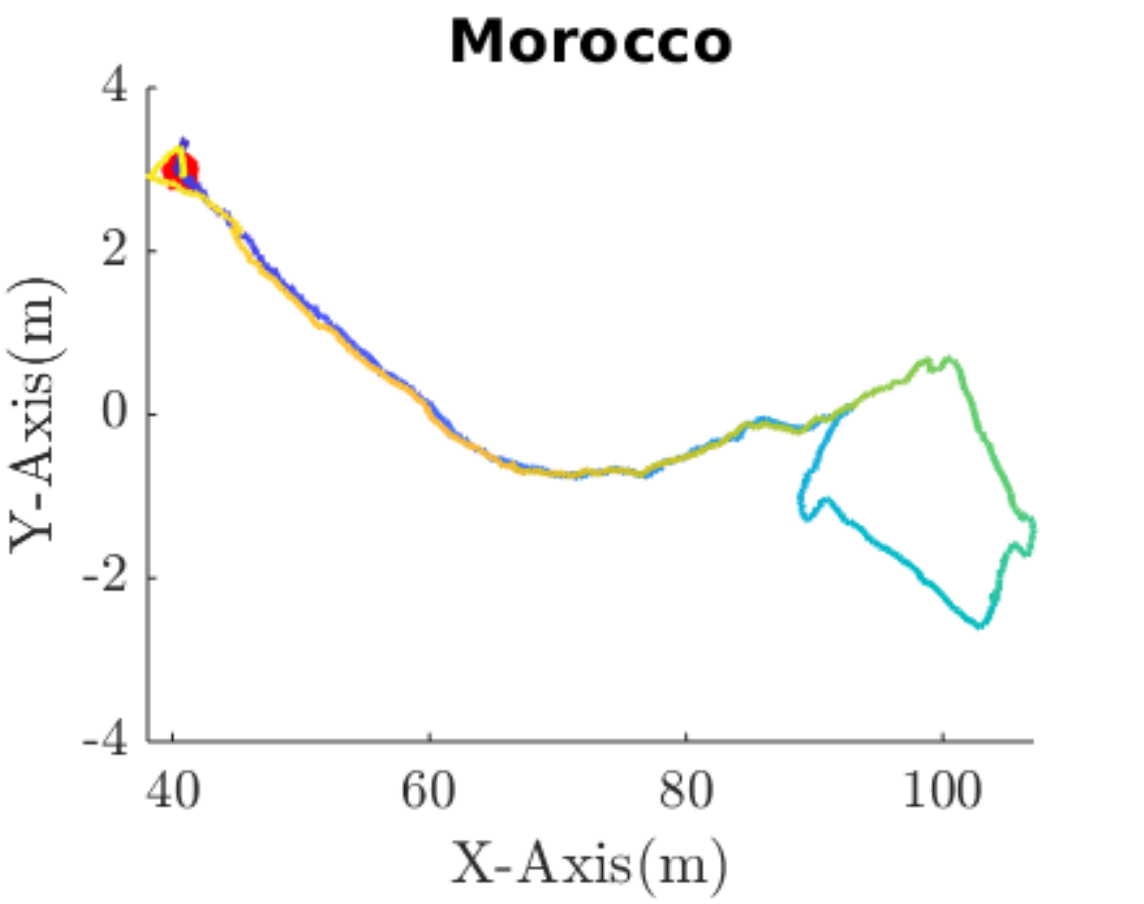}}
	\caption{Ground-truth trajectories of the datasets computed using DGPS data (not shown for Etna3 due to missing data). Trajectories are colored according to the timestamps. The start positions are shown with a red circle and the end positions with a red square.}
	\label{fig:dgps_tracks}
\end{figure}

The  pipeline was tested with five datasets.
Four were collected with the Lightweight Rover Unit (LRU) (Fig~\ref{fig:lru}) during the 2017 Etna demo space mission under the scope of the ROBEX project~\cite{Wedler2017FirstRO}.
The LRU is a lightweight space rover prototype equipped with an IMU, wheel odometry and a stereo camera as the primary sensor for mapping.
A detailed description of the LRU with its hardware and software architecture can be found in~\cite{Schuster2017TowardsAP}.\\
The fifth dataset used has been collected in a planetary analogue environment in Morocco in 2018.
The hand-held sensor suite employed (same as the one integrated into the rover) is shown in Fig.~\ref{fig:first}.
This dataset has been specifically collected to validate mapping and navigation algorithms in particularly challenging unstructured environments.
Different datasets collected in Morocco are in the process of being made publicly available for the robotics community.\\
From hereon we refer to the five datasets as Etna1, Etna2, Etna3, Etna4 and Morocco.
To give an impression of the trajectories from these datasets, ground-truth position tracks as computed from DGPS are shown in Fig.~\ref{fig:dgps_tracks}. To give an impression of views from the navigation camera instead, see Fig.~\ref{fig:third}, and Fig.~\ref{fig:fourth}.

\subsection{Evaluation metrics}

We evaluate the performances of our loop closure detector using the \textit{precision-recall} metric:
\begin{equation}
\text{P} = \frac{\text{T}_p}{\text{T}_p + \text{F}_p} \quad \text{R} = \frac{\text{T}_p}{\text{T}_p + \text{F}_n},
\end{equation}
where $\text{T}_p$ is the number of correct submap matches validated by the proposed method, $\text{F}_p$ is the number of wrong matches, and $\text{F}_n$ is the number of true submap matches which were not detected.
We choose to determine whether a match is true or false from the spatial overlap of submaps, which are positioned in world coordinates using the estimated poses from our SLAM system \cite{doi:10.1002/rob.21812}.
We consider the output of SLAM accurate enough to validate the approach.
Specifically, we compute a bounding box for each submap in the $x$ and $y$ global coordinates.
Let $\text{A}({S^m})$ and $\text{A}({S^n})$ be areas of the bounding boxes for submaps $S^m$ and $S^n$.
The overlap $o(S^m, S^n)$ is defined as the 
Intersection over Union (IoU):
\begin{equation}
o(S^m, S^n) = \frac{\text{A}({S^m}) \cap \text{A}({S^n})}{\text{A}({S^m}) \cup \text{A}({S^n})} \quad \in[0, 1].
\end{equation}
Resulting from empirical considerations about our data, for the Etna datasets we decide that true matches are those for which $o(S^m, S^n)>0.3$, and for Morocco where we choose a minimum overlap of 0.1.
Although it is guaranteed that submaps which have little to none overlap are false matches, overlapping submaps which have no unique 3D structure will not be matched regardless of the parameter choices for our algorithm.
Therefore, performances will be penalized in terms of \textit{recall} in most datasets.
Precision-recall curves are generated by varying the minimum number of inlier SURF matches $n$ between GP gradient maps from 1 to 20.

\subsection{The inadequacy of visual-only approaches}
\begin{figure}
	\centering
	\includegraphics[width=\linewidth]{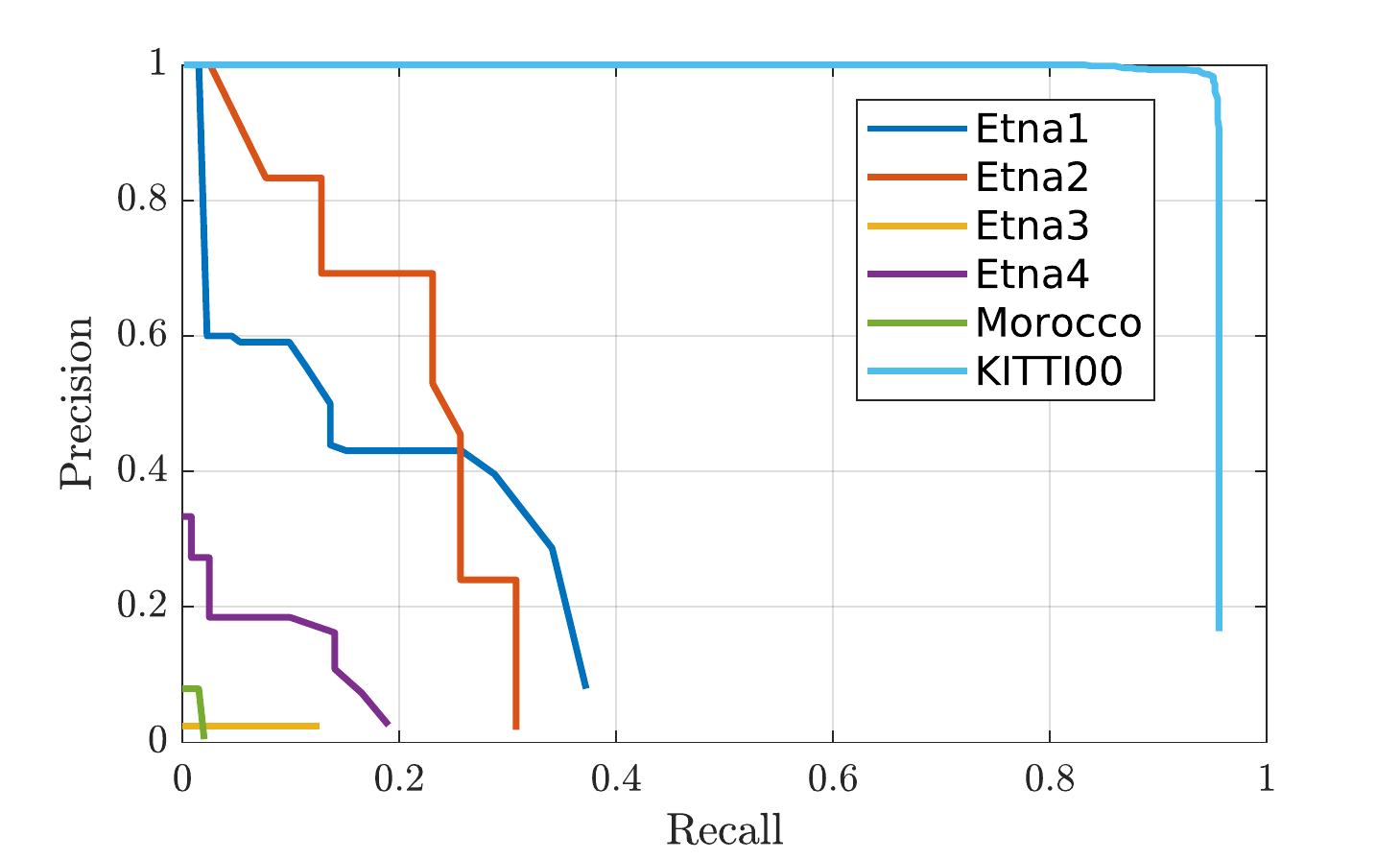}
	\caption{Results for place recognition with the state-of-the-art, visual-only iBoW-LCD approach \cite{Garcia-Fidalgo2018}. We compare precision-recall curves obtained from the KITTI autonomous driving dataset with those obtained from our data. We conclude that traditional visual place recognition is perfectly adequate in the context of autonomous driving, but the performance drops significantly on images captured by a planetary exploration rover in challenging unstructured environments.}
	\label{fig:visualPR}
\end{figure}

Traditional loop closure detection based on image features and bag-of-words demonstrates excellent performances in datasets oriented towards autonomous driving.
Nonetheless, the extreme visual aliasing and strong viewpoint differences encountered in planetary exploration scenarios pose great challenges for visual-only place recognition algorithms.
To demonstrate this, we evaluated the state-of-the-art incremental loop-closure detector iBoW-LCD \cite{Garcia-Fidalgo2018} on images from our Etna and Morocco datasets as well as on the first sequence from the KITTI dataset \cite{Geiger2012}.
We extracted 1500 ORB features per frame and adapted the parameters recommended in the paper to achieve the best performances in each dataset.
True matches were labelled based on the spatial and viewpoint proximity of the camera pose related to each frame.
Fig~\ref{fig:visualPR} shows the precision-recall curves obtained with iBoW-LCD applied to our datasets and the KITTI00 sequence.
The appearance-based place recognition method performs well on the KITTI sequence thanks to the specific motion of the car and the distinctive features present in man-built environments.
However, iBoW-LCD performed poorly on the Etna and Morocco datasets due to the ambiguous visual appearance and the proximity of the viewpoints with the ground.
Furthermore, on the Morocco sequence, visual-only loop-closure detection through standard appearance approaches is not possible as the camera trajectory repeats the same path from an almost opposite direction.
This experiment demonstrates the necessity for alternative solutions.
In the following subsection, we will show how our proposed approach based on GP gradient maps helps to overcome the limitations of traditional place recognition methods.

\subsection{GP gradient maps vs. 3D feature matching}
Here we evaluate the proposed pipeline for loop closure detection against a traditional approach based on matching 3D features extracted from submap point clouds. To this end,
keypoints are sampled from high curvature regions as in~\cite{8950110} and matched across submap pairs using SHOT and CSHOT features.
The resulting keypoint matches are then filtered from outliers using a RANSAC approach as implemented in the PCL library~\cite{Rusu2011}, setting the probability of selecting an outlier-free model to 0.99.

\begin{figure}
	\centering
	\subfloat[Etna1]{\includegraphics[width=0.49\columnwidth]{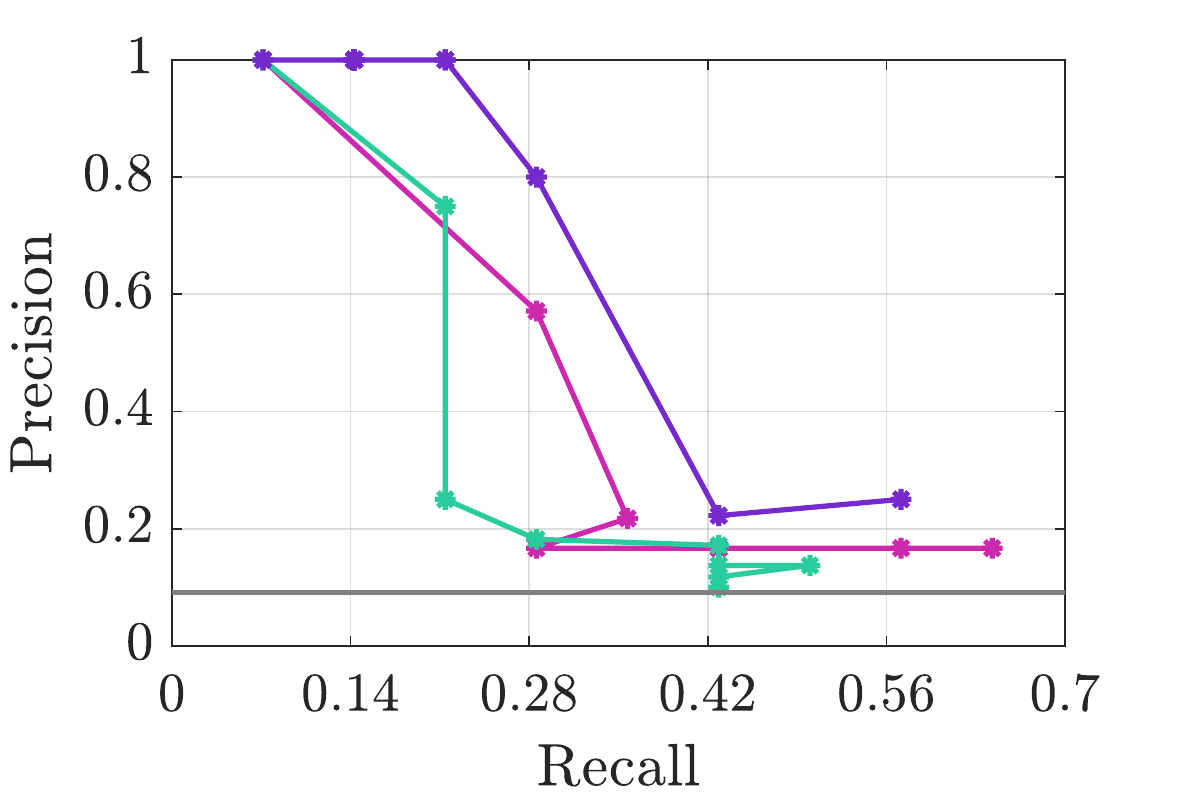}}
    \hfill
    \subfloat[Etna2]{\includegraphics[width=0.49\columnwidth]{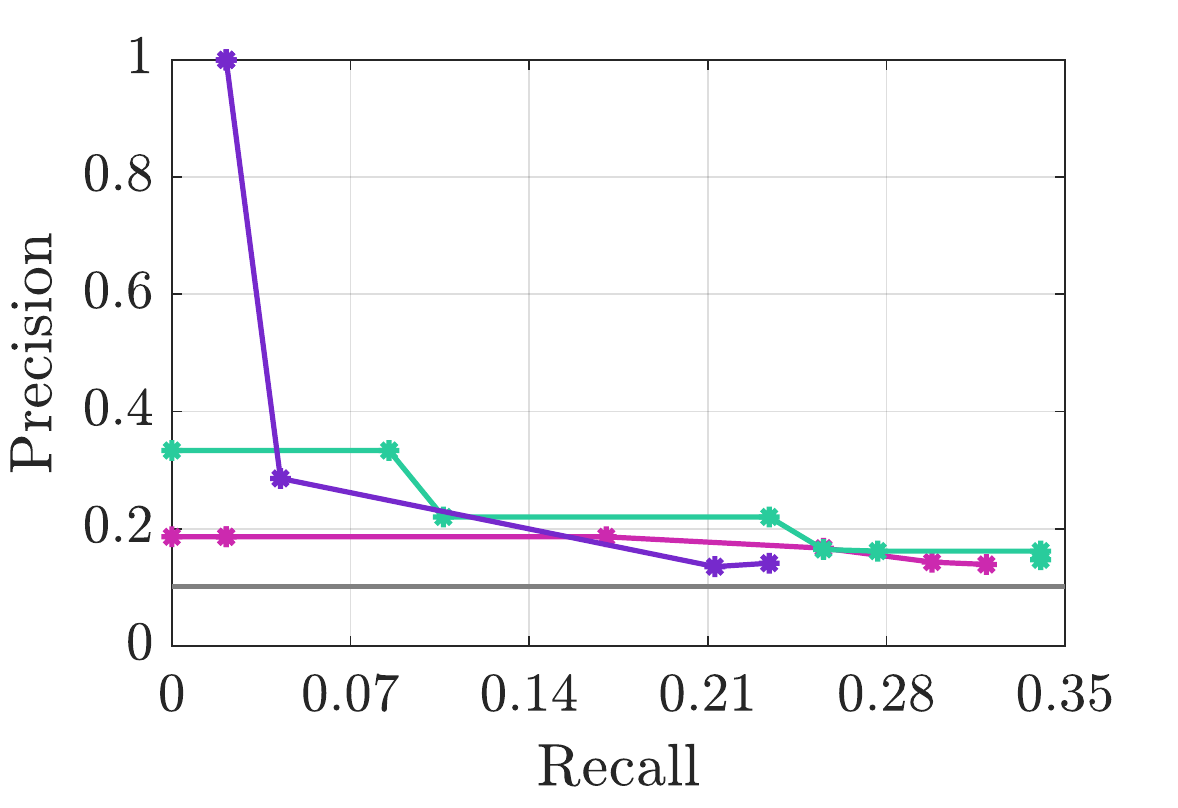}}
    \\
	\subfloat[Etna3]{\includegraphics[width=0.49\columnwidth]{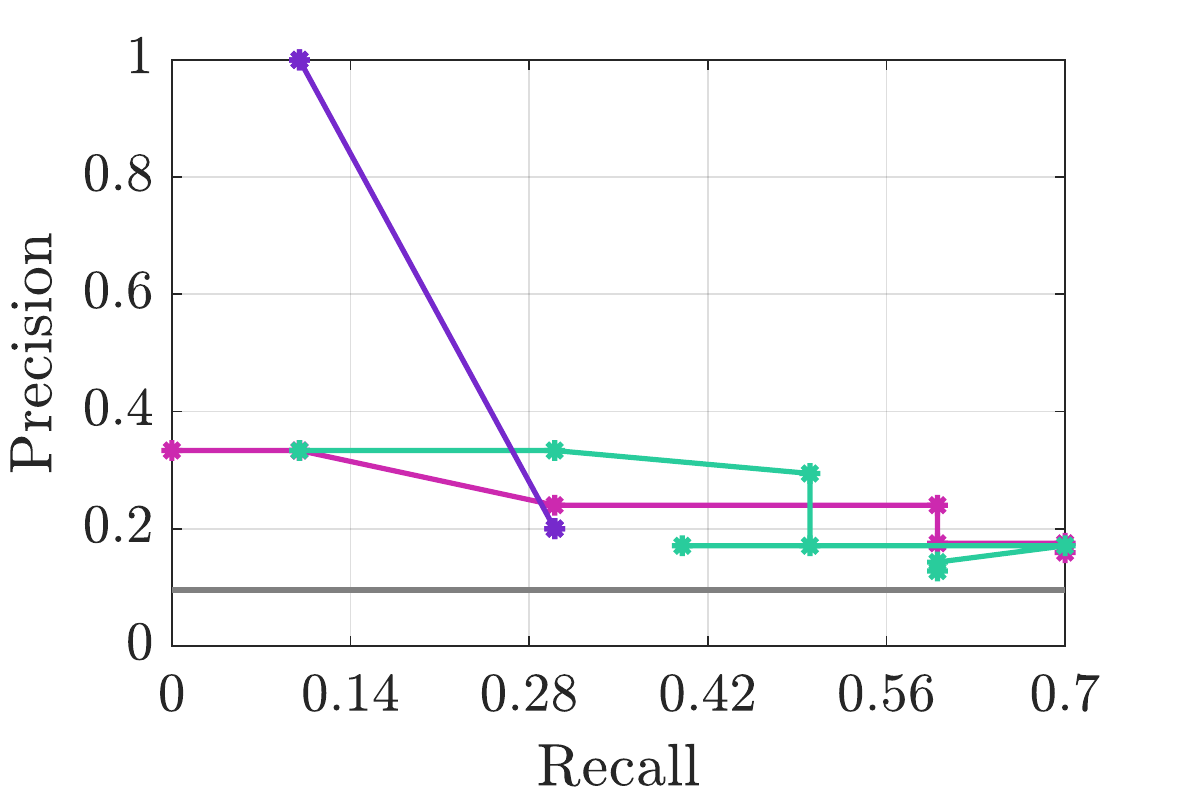}}
    \hfill
    \subfloat[Etna4]{\includegraphics[width=0.49\columnwidth]{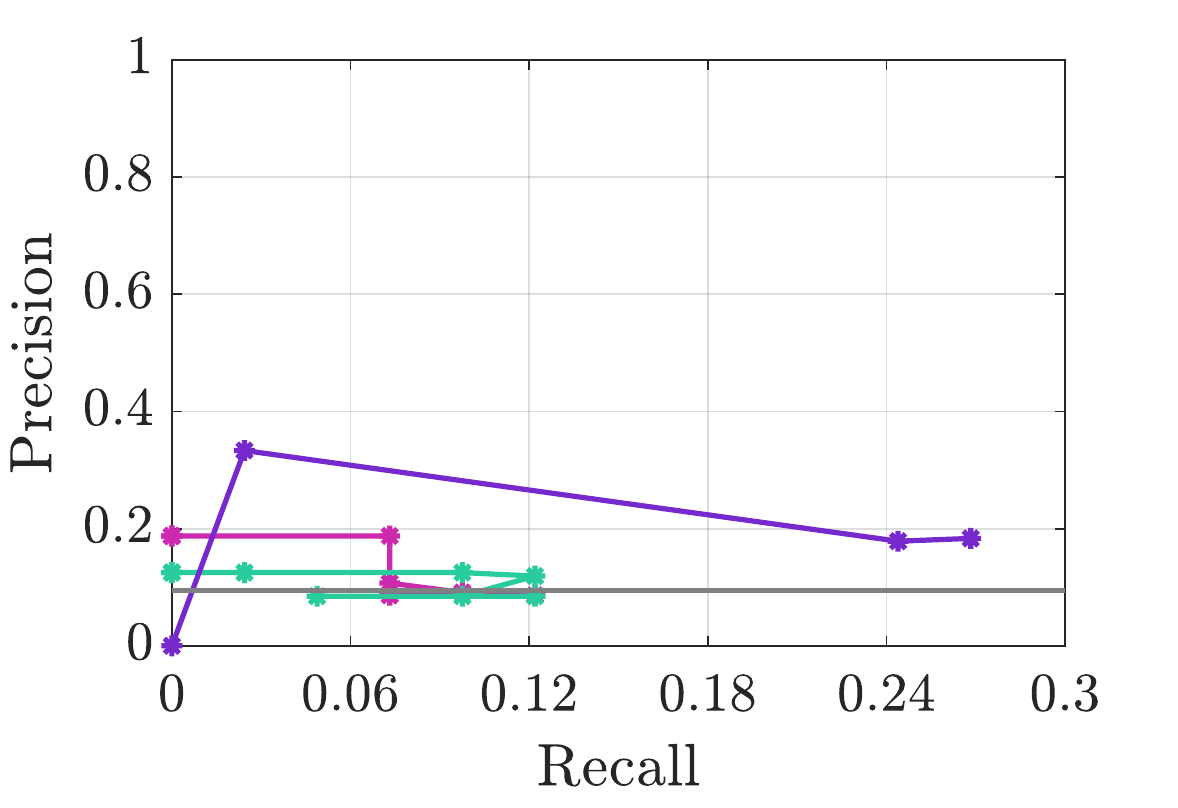}\label{fig:PR_etna_4}}
    \\
	\subfloat[Morocco]{\includegraphics[width=0.49\columnwidth]{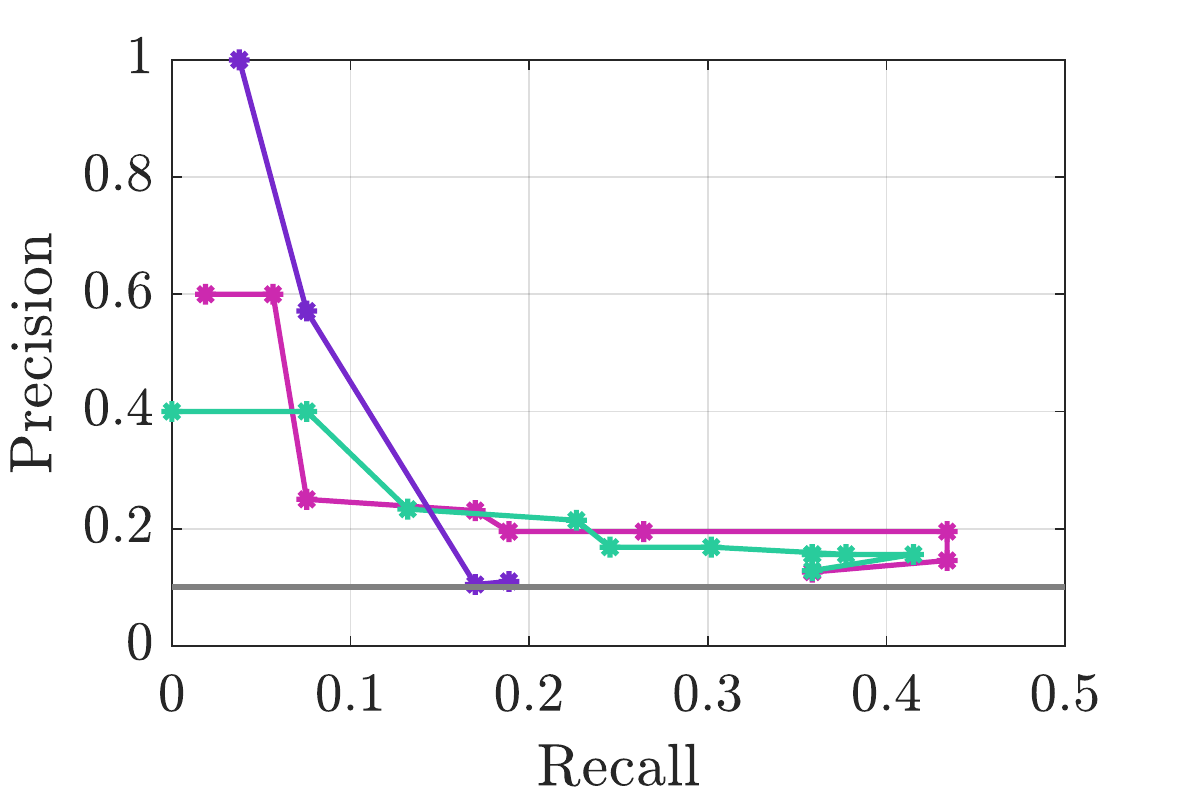}}
    \hfill
    \subfloat{\includegraphics[width=0.44\columnwidth]{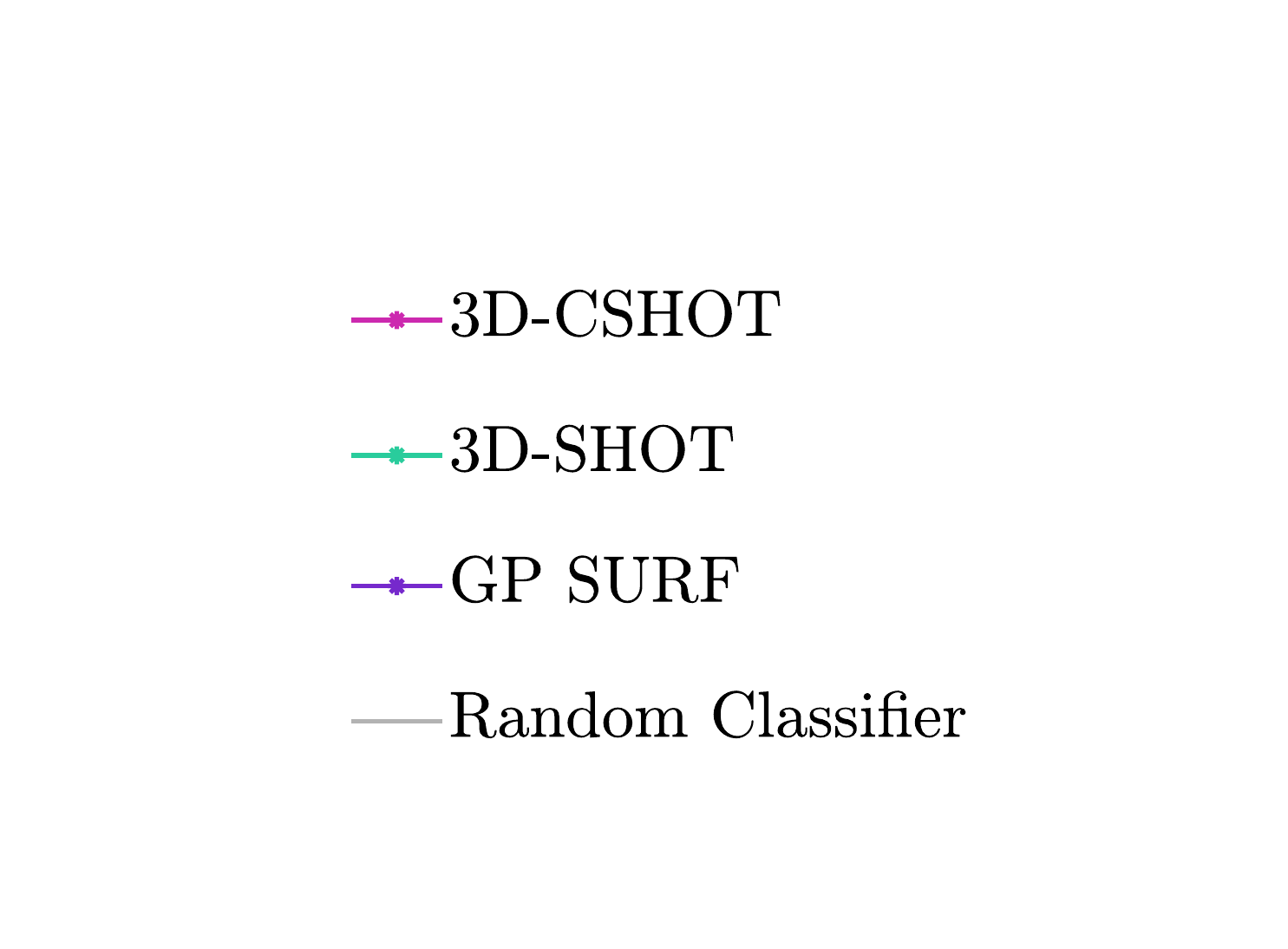}}
	\caption{Precision-recall curves comparing the performance of the proposed GP gradient method with 3D descriptor matching as well as random classifier.}
	\label{fig:pr_curves_inliers}
\end{figure}

\begin{figure}
\centering
    \includegraphics[trim=1cm 0 1cm 0, clip, width = \columnwidth]{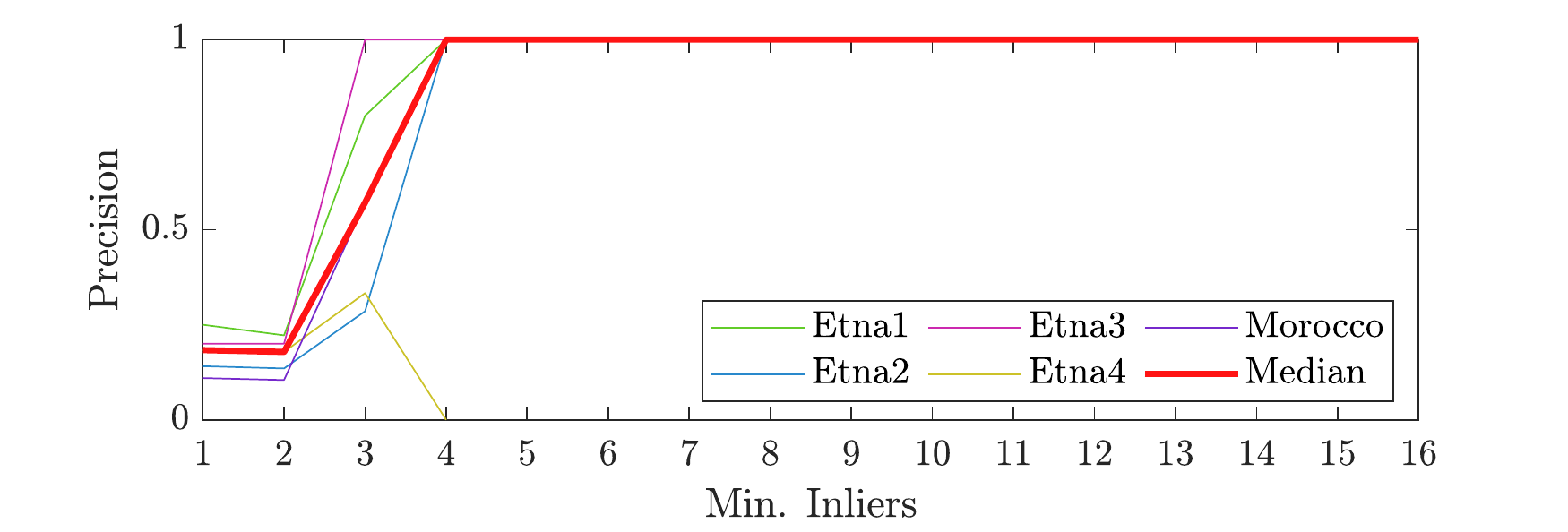}
    \caption{Average precision across all the datasets as a function of the threshold on the number of inliers in the proposed RANSAC-based matching.}
\label{fig:prec_vs_inl}
\end{figure}

Fig.~\ref{fig:pr_curves_inliers} shows the precision-recall curves for the Etna and Morocco sequences.
As can be observed for all tests, except for Etna4 in Fig.~\ref{fig:PR_etna_4}, all the curves for our method start at 100\% precision.
The curves show that our approach is more accurate and conservative than the baseline:
for a set of input parameters, the gradient maps pairs that pass the RANSAC test for SURF matching will likely be all true,
while many validated matches from CSHOT+RANSAC and SHOT+RANSAC are wrong even for the more conservative descriptor matching thresholds.
Fig.~\ref{fig:prec_vs_inl} reports the precision obtained in all datasets as a function of the minimum number of inlier SURF matches after RANSAC $n_\text{inl}$.
The plot shows that for $n_\text{inl}\geq4$, submap matches can be selected with 100\% precision on all datasets except for Etna4, which the precision-recall curves in Fig.~\ref{fig:prec_vs_inl} prove to be a very challenging scenario for loop closure detection.
Indeed on Etna4 performances are very limited also for appearance-based methods, see Fig.~\ref{fig:visualPR}. For $n_\text{inl} \geq 5$ all the selected matches are instead correct, achieving the highest precision among the tested baseline approaches.

\begin{figure}
\centering
    \includegraphics[width=0.95\columnwidth, clip, trim=0cm 5cm 0cm 6cm]{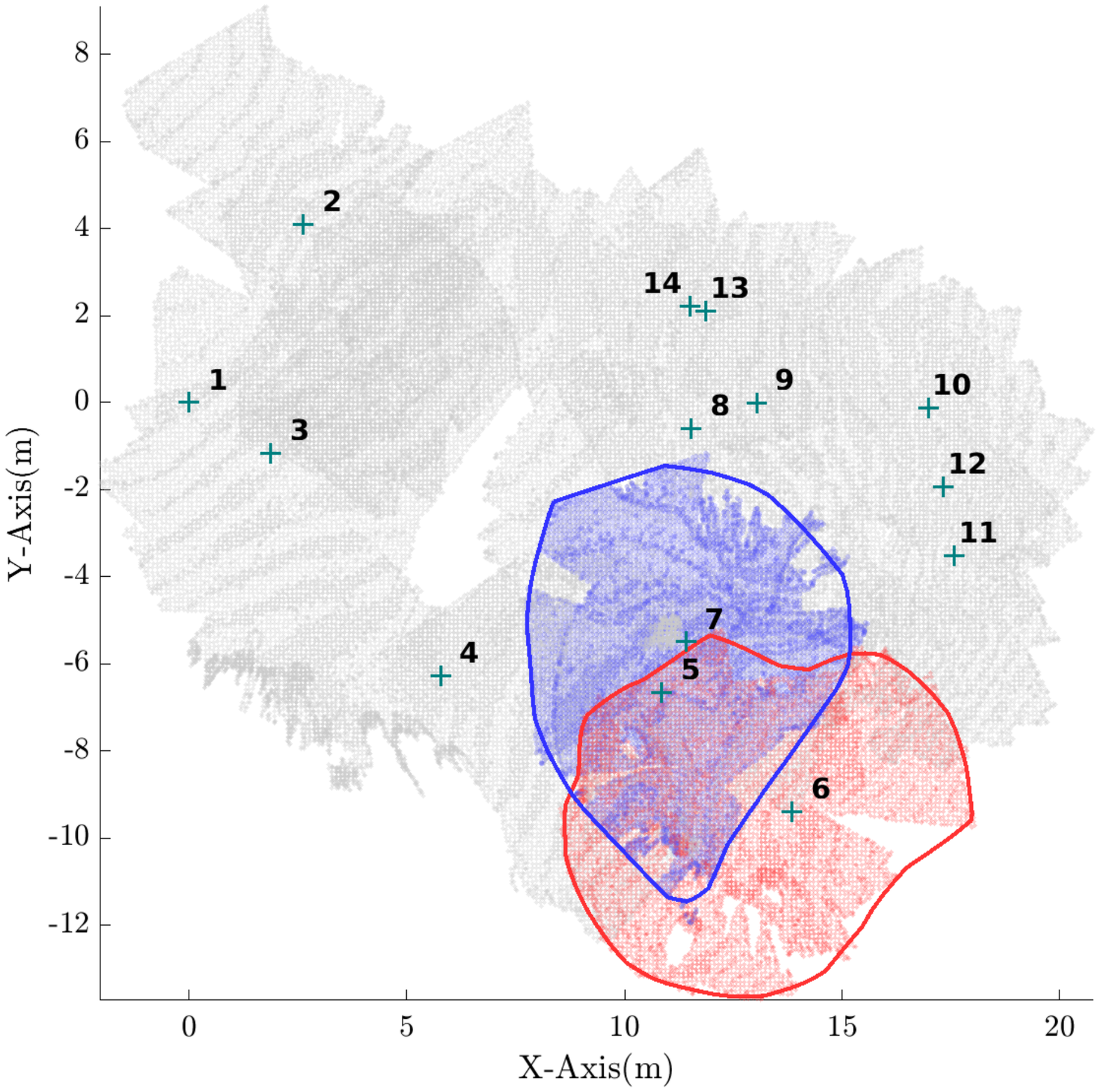}
\caption{The two identified loop closure candidates from Etna3 i.e. submaps 5 and 6 in red and blue, rest of the submaps from the session in grey aligned using pose estimation results. The numbers represent the submap index and the position of each submap local reference frame estimated by the visual-inertial odometry is marked with a '+' symbol.}
\label{fig:alignedClouds}
\end{figure}

The two matching submaps that were correctly identified from Etna3 with this method are shown in Fig.~\ref{fig:alignedClouds} as an example. Notice how for Etna3, loop closures can not be established with the tested appearance-based method \cite{Garcia-Fidalgo2018}, see Fig.~\ref{fig:visualPR}. The benefits provided by our method against both 3D feature matching and visual methods especially emerge when the camera traverses a path from opposite directions, such as for the traverse in Morocco (see Fig.\ref{fig:dgps_tracks}). In this case, overlapping submaps
do not share significative structures that can be matched, such as rocks, as they are reconstructed on opposing viewpoints.
In addition, images captured from close positions observe different portions of the environment. Our approach is instead able to deal with missing observations and produce correct submap matches.

\section{Conclusions}\label{sec:conclusions}
In this paper, we presented a method to detect loop closures in unstructured planetary environments based on Gaussian gradient maps.
The novel Gaussian gradient representation is a continuous probabilistic representation of the environment's terrain elevation based on GP regression and the application of linear operators on the kernel.
This method allows image-like inferences of the terrain's elevation alongside its associated variance.
The gradient maps are then matched through a $SE(2)$-constrained feature-based RANSAC algorithm that leverages the uncertainty knowledge of the terrain.
The method and its robustness to noisy input data have been validated in five challenging datasets outperforming traditional 3D feature matching.

Future works include the integration of this pipeline into a real-time multi-modal and multi-session mapping framework leveraging both visual and structural similarities to better generalize to a wider range of scenarios, as well as extending it to heterogeneous sensor setups (LiDAR, stereo and RGB-D cameras).
To this end, work will also be conducted to reduce the computational cost of the overall pipeline.
At the method level, one can easily consider inducing points as in~\cite{Quinonero-Candela2005} to breakdown the cubic complexity of the GP gradient map generation.
The use of bag-of-words techniques like~\cite{Garcia-Fidalgo2018} can greatly reduce the number of RANSAC matching attempts performed.
In terms of implementation, many operations are highly parallelizable (GP inferences, RANSAC runs, matching score, etc.) and would benefit from optimized computation on GPU.

\section*{ACKNOWLEDGMENT}
This work was supported by the UA-DAAD 2018 Australia-Germany Joint Research Cooperation Scheme, project COSMA (contract number 57446007) and the project ARCHES (contract number ZT-0033).

\bibliographystyle{IEEEtran}
\bibliography{bibliography.bib,bibliography_cedric}

\addtolength{\textheight}{-12cm}   







\end{document}